\documentclass{IEEEoj}

\usepackage{cite}
\usepackage{amsmath,amssymb,amsfonts}
\usepackage{algorithmic}
\usepackage{graphicx,color}
\usepackage{textcomp}

\usepackage{xltabular}
\usepackage{booktabs}
\usepackage[colorinlistoftodos, textsize=footnotesize, textwidth=35mm]{todonotes}
\usepackage[nolist]{acronym} 
\usepackage{hyperref} 
\usepackage{harveyballs}
\usepackage{multirow}
\usepackage{rotating}
\usepackage{multirow}
\usepackage{makecell}
\usepackage{duckuments}
\usepackage{float}
\usepackage{dblfloatfix}
\usepackage{ulem}

\usepackage{epsfig}
\usepackage{subcaption}
\usepackage{calc}
\usepackage{amstext}
\usepackage{multicol}
\usepackage{pslatex}
\usepackage{algorithm2e}
\usepackage{siunitx}
\usepackage{colortbl}
\usepackage{array}
\usepackage[nameinlink, capitalise]{cleveref}
\usepackage{pgfplots}
\usepackage{adjustbox}

\newcolumntype{P}[1]{>{\raggedleft\arraybackslash\linespread{0.8}\selectfont}p{#1}}
\pgfplotsset{ every non boxed x axis/.append style={x axis line style=-},
     every non boxed y axis/.append style={y axis line style=-}}
\pgfplotsset{xtick style={draw=none}}
     
\definecolor{1}{HTML}{0065BD}
\definecolor{2}{HTML}{4590D1}
\definecolor{3}{HTML}{6DA9DC}
\definecolor{4}{HTML}{95C2E8}
\definecolor{5}{HTML}{BDDAF4}
\definecolor{6}{HTML}{E5F3FF}

\def\BibTeX{{\rm B\kern-.05em{\sc i\kern-.025em b}\kern-.08em
    T\kern-.1667em\lower.7ex\hbox{E}\kern-.125emX}}
\AtBeginDocument{\definecolor{ojcolor}{cmyk}{0.93,0.59,0.15,0.02}}
\def\OJlogo{\vspace{-14pt}\includegraphics[height=28pt]{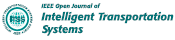}}

\begin{document}
\receiveddate{XX Month, XXXX}
\reviseddate{XX Month, XXXX}
\accepteddate{XX Month, XXXX}
\publisheddate{XX Month, XXXX}
\currentdate{XX Month, XXXX}
\doiinfo{OJITS.2022.1234567}

\title{Evaluation of Teleoperation Concepts to solve Automated Vehicle Disengagements}

\author{David Brecht\textsuperscript{1}, 
        Nils Gehrke\textsuperscript{1}, 
        Tobias Kerbl\textsuperscript{1},
        Niklas Krauss\textsuperscript{1},
        Domagoj Majstorovi\'c\textsuperscript{1},
        Florian Pfab\textsuperscript{1},
        Maria-Magdalena Wolf\textsuperscript{1} and 
        Frank Diermeyer\textsuperscript{1}}

\affil{Institute of Automotive Technology, Technical University of Munich (TUM), Boltzmannstr. 15, 85748 Garching b. M\"unchen, Germany}
\corresp{CORRESPONDING AUTHOR: David Brecht (e-mail: david.brecht@tum.de).}
\authornote{This work was supported by the Federal Ministry of Education and Research of Germany~(BMBF) within the project AUTOtech.\textit{agil}~(FKZ~01IS22088), the project Wies'n Shuttle~(FKZ~03ZU1105AA) in the MCube cluster, the Federal Ministry of Economic Affairs and Climate Actions within the project Safestream~(FKZ~01ME21007B), the project ATLAS-L4~(FKZ~19A21048l) and through basic research funds from the Institute for Automotive Technology.}

\markboth{Preparation of Papers for IEEE OPEN JOURNALS}{Author \textit{et al.}}

\begin{abstract}
Teleoperation is a popular solution to remotely support highly automated vehicles through a human remote operator whenever a disengagement of the automated driving system is present.
The remote operator wirelessly connects to the vehicle and solves the disengagement through support or substitution of automated driving functions and therefore enables the vehicle to resume automation. 
There are different approaches to support automated driving functions on various levels, commonly known as teleoperation concepts.
A variety of teleoperation concepts is described in the literature, yet there has been no comprehensive and structured comparison of these concepts, and it is not clear what subset of teleoperation concepts is suitable to enable safe and efficient remote support of highly automated vehicles in a broad spectrum of disengagements.
The following work establishes a basis for comparing teleoperation concepts through a literature overview on automated vehicle disengagements and on already conducted studies on the comparison of teleoperation concepts and metrics used to evaluate teleoperation performance.
An evaluation of the teleoperation concepts is carried out in an expert workshop, comparing different teleoperation concepts using a selection of automated vehicle disengagement scenarios and metrics.
Based on the workshop results, a set of teleoperation concepts is derived that can be used to address a wide variety of automated vehicle disengagements in a safe and efficient way.
\end{abstract}


\maketitle

\section{\MakeUppercase{Introduction}}
\label{sec:introduction}

\acp{av} are one of the key technologies of tomorrow's mobility, as they contribute to safer and more efficient mobility systems. %
Their development has been a focus of academic and industrial research for decades, and market introduction is increasingly imminent.
The first driverless vehicles without safety drivers can already be found on public roads in various parts of the world, such as Cruise and Waymo in the USA or Baidu in China. 
However, remaining technological challenges such as complex or unknown edge cases pose a challenge to the utilization of \acp{av}. Currently, this threat can be observed at Cruise, where several safety incidents have led to a temporary suspension of the company's driverless services \cite{cruise2023}. 

To address these challenges and accelerate market introduction of \acp{av}, teleoperation is being developed as a fallback solution for Automated Driving functions.
Teleoperation enables a human remote operator, in the following referred to as operator, who is not located in the vehicle to support the \ac{av}. 
Requesting remote support whenever an Automated Driving System disengages or reaches the limits of its \ac{odd}, teleoperation aims to provide a safe and efficient solution to overcome limitations posed by edge cases and resulting disengagements. 
Once the \ac{av} is brought back into its nominal \ac{odd} by the operator it can again continue its journey fully automated as before.

In recent years, a variety of different \acp{tc} relating to how a human operator can remotely support or substitute an Automated Driving function have been proposed in the literature \cite{Maj2022b}.
All of these \acp{tc} were either designed to support the \ac{av} in specific scenarios or to address teleoperation challenges such as latency or limited situational awareness.
Yet, existing literature does not provide a complete comparison of the \acp{tc} considering their ability to support \acp{av}.
This gap in the literature is addressed in this paper using a methodological approach. 

\subsection{Taxonomy}
\label{subsec:Taxonomy}

With the goal of introducing a standardized taxonomy and facilitating a precise formulation of the problem for the following work, this subsection provides a breakdown of the teleoperation system into three modules and introduces the associated tasks for the operator.

\begin{figure}[H]
  \centering
  \includegraphics[width = 8cm]{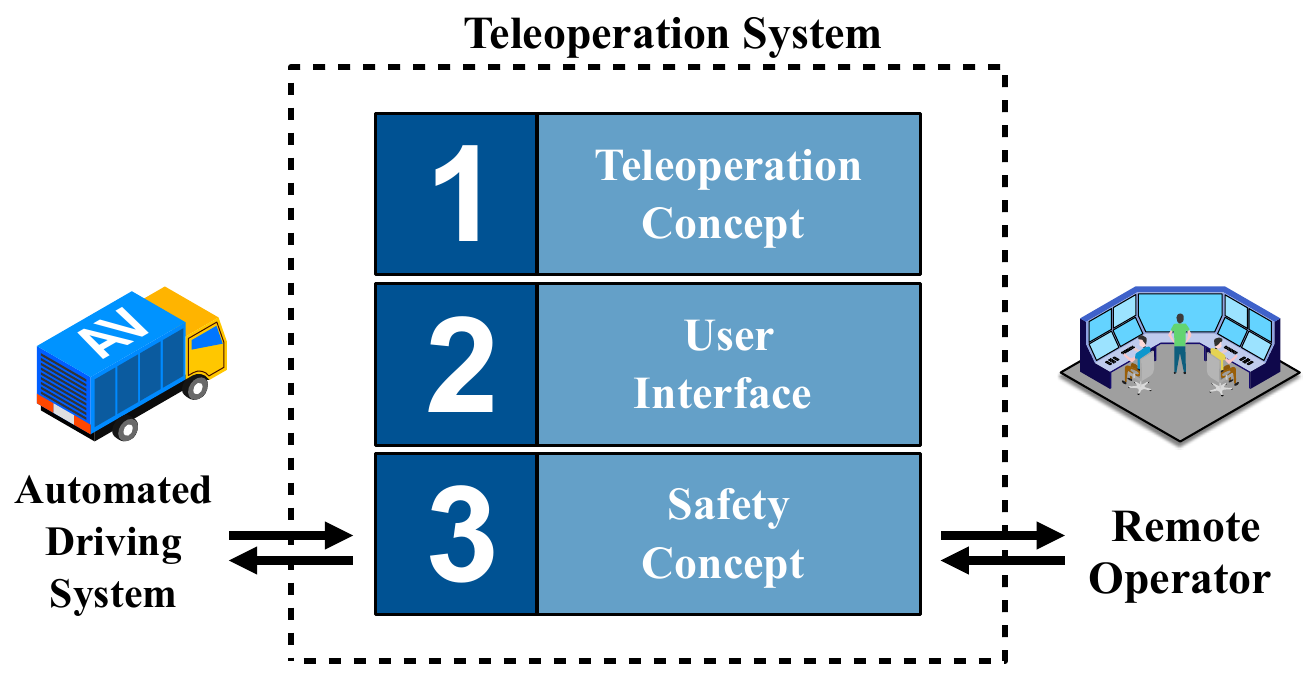}
  \caption{A teleoperation system comprises a teleoperation concept, a user interface and a safety concept}
  \label{fig:teleopSystem}
\end{figure}

As shown in \cref{fig:teleopSystem}, a teleoperation system comprises three independent components, namely \ac{tc}, user interface and safety concept. 
\acp{tc}, which are the focus of the present work, describe the interaction type by which the operator supports the automation of the vehicle. 
This can be direct control inputs like steering wheel angle and velocity command, trajectories or high-level instructions addressing specific Automated Driving functions.
The user interface describes the human-machine interface on the operator side.
The safety concept includes additional safety approaches such as an emergency brake assist or network monitoring which serve the goal of increasing system safety. 

Literature reviews for vehicle teleoperation display a variety of possible teleoperation tasks \cite{Maj2022b,Bog2022b,Ket2021,Ket2021b}. 
The following categorization aims at contributing to a more precise task definition.

\textit{Driving-related teleoperation tasks} include all tasks required to maneuver a vehicle on public roads in conformity with respective legislation.
In the context of \ac{av} teleoperation, this implies the direct interaction with or the replacement of Automated Driving functions that are part of the Dynamic Driving Task.
This includes setting the velocity, giving waypoints, modifying the \ac{odd} or the environmental model, using the indicator, and assessing the current system performance. These tasks are independent of the specific vehicle, its mission, and its passengers or cargo.

\textit{Mission-related teleoperation tasks} on the other hand describe additional tasks necessary for successful mission completion. 
This includes, e.~g., communication with passengers in the vehicle, checking door obstructions, manual vehicle rerouting or adapting air conditioning.
In the present work, the focus is on driving-related teleoperation tasks.

\subsection{Contributions}
The objective of the following work is to identify a set of \acp{tc} that is needed to solve a wide variety of \ac{av} disengagements in a safe and efficient manner.
It can therefore be used to universally support \acp{av}.
This reduction to a minimum set of concepts reduces the complexity of the system implementation and
highlights which concepts are worth further research.

To derive the concept set in a structured manner, this paper first presents the results of a literature analysis on \ac{av} disengagements and on existing comparisons of individual \acp{tc} in \cref{sec:avfails} and 
\cref{sec:comparison}, respectively.
Based on these prerequisites, an expert workshop is conducted where the concepts are compared against a selection of scenarios and \acp{kpi} from the previous sections.
The expert workshop aims to fill the literature gaps regarding the comparison of \acp{tc}.
The procedure and results of the workshop are described in \cref{sec:workshop}.
Finally, in \cref{sec:concept} a holistic teleoperation set of concepts is derived that is able to address a wide range of \ac{av} disengagements in a safe and efficient manner.
The limitations of the results are discussed, and potential future work is pointed out in \cref{sec:conclusion}.

\section{\MakeUppercase{Why do Automated Vehicles fail?}}
\label{sec:avfails}
\begin{table*}[bh]
\small %
\centering
\caption{Overview on the distribution and sources for the \ac{av} fail causes \cite{Zha2022}\cite{Rich2023} and the related exemplary scenarios from the literature. Due to different degrees of detail, the distribution of \cite{Zha2022} contains different categories names if a direct match was not possible.}
\begin{tabular}{lcccl}
     \toprule
     \makecell[l]{Fail Cause \cite{Rich2023}}
     & \makecell{Solvable by \\ teleoperation}   
     & \makecell{Richter et al. \cite{Rich2023} \\ CA DMV report \\2021} 
     & \makecell{Zhang et al. \cite{Zha2022} \\ CA DMV reports \\2014 - 2020} 
     & \makecell[l]{Example Scenarios}  \\ 
     \midrule
     Object detection 
     & Yes 
     & 708 
     & 1320 (Perception) 
     & Difficult weather \cite{Ket2022}, lighting-induced false positives \cite{Ten2023}  \\ 
     Path planning 
     & Yes 
     & 669 
     & 7507 
     & Vehicle parked in second row \cite{Ket2022}, Puddles \cite{Ten2023} \\
     Localization 
     & Yes 
     & 294 
     & 2198 (Planning) 
     & State-Estimation-Failure \cite{Ket2022}\\
     Trajectory planning 
     & Yes 
     & 254 
     & 7560 
     & Complex traffic scenarios at intersections \cite{Ten2023} \\
     Lane detection 
     & Yes 
     & 157 
     & 1320 (Perception) 
     & Unmapped construction site \cite{Ket2022}, lane marking issues \cite{Ten2023} \\ 
     \makecell[lt]{Actuators\\ } 
     & No 
     & 137 
     & 2362 (Control) \\
     Software Components 
     & No 
     & 135 
     & 989 (System general) \\
     \midrule
     Total disengagements 
     & 
     & 2629 
     & 9511 
     & \\
     \bottomrule
\end{tabular}
\label{tab:av_fail_scenario}
\end{table*}

The necessity to assist or hand over the Dynamic Driving Task arises from potential \ac{av} disengagements meaning a deactivation of the Automated Driving System \cite{ccr_37}. 
Reasons for such disengagements may include failures within the Automated Driving System or when an \ac{odd} boundary is reached. 
In case of driverless operation of Level~4 or Level~5 \acp{av}, according to the standard J3016 \cite{SAEJ3016} of the \ac{sae}, the vehicles are required to reach a Minimal Risk Condition before they disengage their Automated Driving System. 
Being in the Minimal Risk Condition, a risk-minimal stopped state, the vehicle can not continue its mission by itself.
For this reason, an operator can, in part or entirely, take over control of the Dynamic Driving Task to address the present system failures or return the \ac{av} to its \ac{odd} \cite{SAEJ3016}. The operator's aim in such situations is to continue the vehicle's mission until the \ac{av} is able to fully perform the Dynamic Driving Task again.

To derive suitable \acp{tc} for future application as a mission fallback for \acp{av}, it is necessary to first analyze the context in which teleoperation is needed.
As stated previously, the operator is required in the case of \ac{av} disengagements to continue the vehicle's mission. 
This implies that the respective teleoperation concepts must be able to solve scenarios which lead to a disengagement and that might occur after it.

For a more detailed understanding of \ac{av} disengagements, common causes for disengagements are outlined in the following. Existing literature is taken into account and based on the state of the art a classification of disengagement scenarios is derived.
Based on the identified causes, only the disengagement scenarios that can theoretically be solved by an operator are considered. 
This means that causes like a technical failure are excluded and not further considered.
In the next step, a subset of specific scenarios that represent each disengagement class is generated. 
The scenarios are later used in the expert workshop as they illustrate a variety of \ac{av} disengagements in a comprehensive way. 

Disengagements from \ac{av} present a valuable source to identify relevant scenarios to evaluate \acp{tc} for \ac{av} support. 
A major source for disengagements in the literature are the disengagement reports published by the California Department of Motor Vehicles \cite{ca_disengagement_reports}. 
Existing analysis of the respective data however notes that despite the large database, specific scenarios can only be extracted to a limited amount \cite{Zha2022, Zhe2023, Sin2021}. Some analysis do not even consider the disengagement cause due to the limited available data in the disengagement reports \cite{Zhe2023, Sin2021}.
Further sources for the scenarios are publications, e.g., from Kettwich~et~al. \cite{Ket2022, Ket2021b} that include scenarios from \ac{av} operation in Germany.

With the disengagements from the California Department of Motor Vehicles being most commonly used in the literature, the \ac{av} disengagements were extracted from analysis of this source. Since the exact scenario that provoked the failure is difficult to extract from the disengagement data, the focus is shifted to the causes of the failure. The respective failure causes and their level of detail however highly depend on the applied method of analysis, as results from different sources\cite{Rich2023, Zha2022, Guo2022, Wu2023}. Further the number of disengagements for each fail cause differ with respect to the selected analysis-method and the time-frame of the considered reports.

For the extraction of relevant scenarios for teleoperation, the latest analysis on \ac{av} disengagements \cite{Rich2023} considering reports between December 2020 and November 2021 was selected. The different fail causes are sorted with respect to the number of related disengagements from high to low, showing that a significant drop of the disengagement number was present after the category of \textit{Software Components} with 135 disengagements \cite{Rich2023}. Further, the disengagements are evaluated based on the possibility to generally solve them with teleoperation. Fail causes where teleoperation can not continue the mission include general software failures or actuator failures. The resulting set of fail causes are thus the most likely to lead to a teleoperation: object detection, lane detection, trajectory planning, and path planning. A detailed list of the categories and corresponding exemplary scenarios is displayed in Tab. \ref{tab:av_fail_scenario}.

With these main causes are selected as disengagement categories, a representation of the fail causes via precise scenarios is required. The majority of disengagements available in the annual reports of the California Department of Motor Vehicles allows only a restricted reconstruction of the scenario resulting in the described disengagement. Thus, disengagement scenario generation must be performed via additional literature sources.

To obtain disengagement scenarios from the selected disengagement categories, the fail causes are mapped to scenarios available in existing teleoperation publications. The performed matching of related scenarios to the fail causes is displayed in Tab. \ref{tab:av_fail_scenario}. 
\begin{figure*}[!t]
    \centering
    \includegraphics[width=1\linewidth]{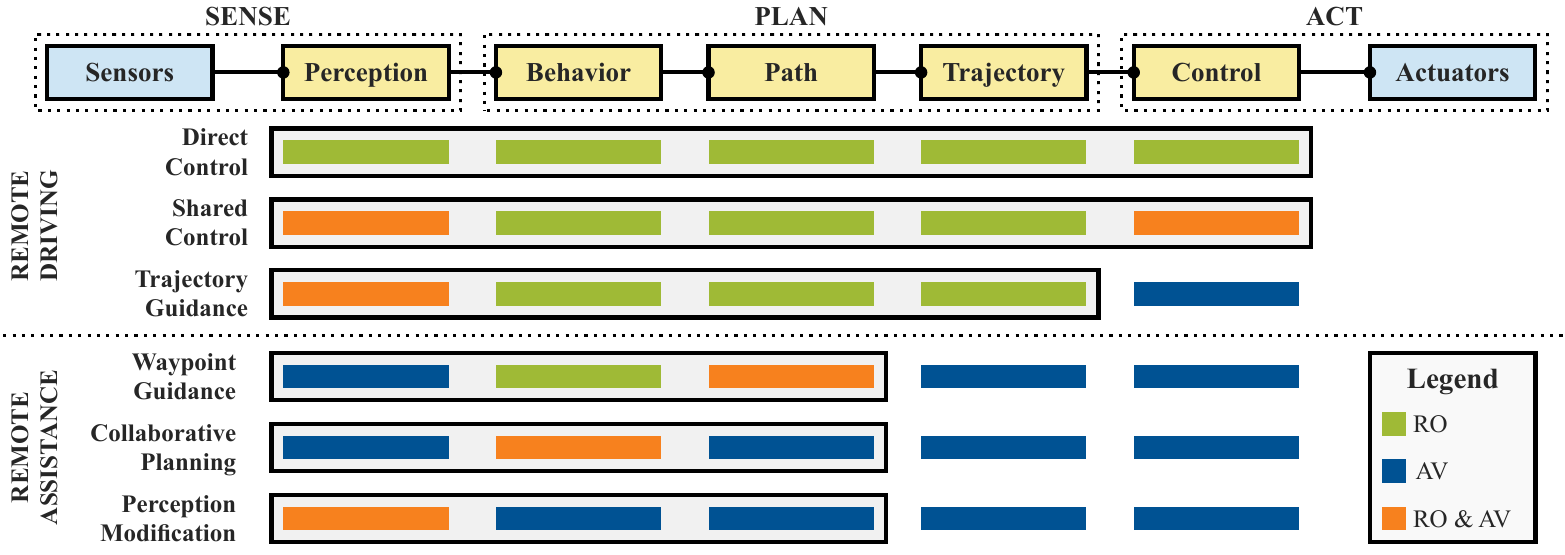}
    \caption{Overview of teleoperation concepts. Colors represent the task division between the remote operator (RO) and the automated vehicle (AV), with bounding boxes in grey depicting the level of module interaction.} \label{fig:concepts}
\end{figure*}

\section{\MakeUppercase{Teleoperation Concepts to solve Automated Vehicle Disengagements}}
\label{sec:comparison}

\begin{figure*}[hbp]
    \centering
    \begin{subfigure}{0.33\textwidth}
        \includegraphics[width=\linewidth, trim={4.5cm 3.5cm 4.5cm 3.5cm}, clip]{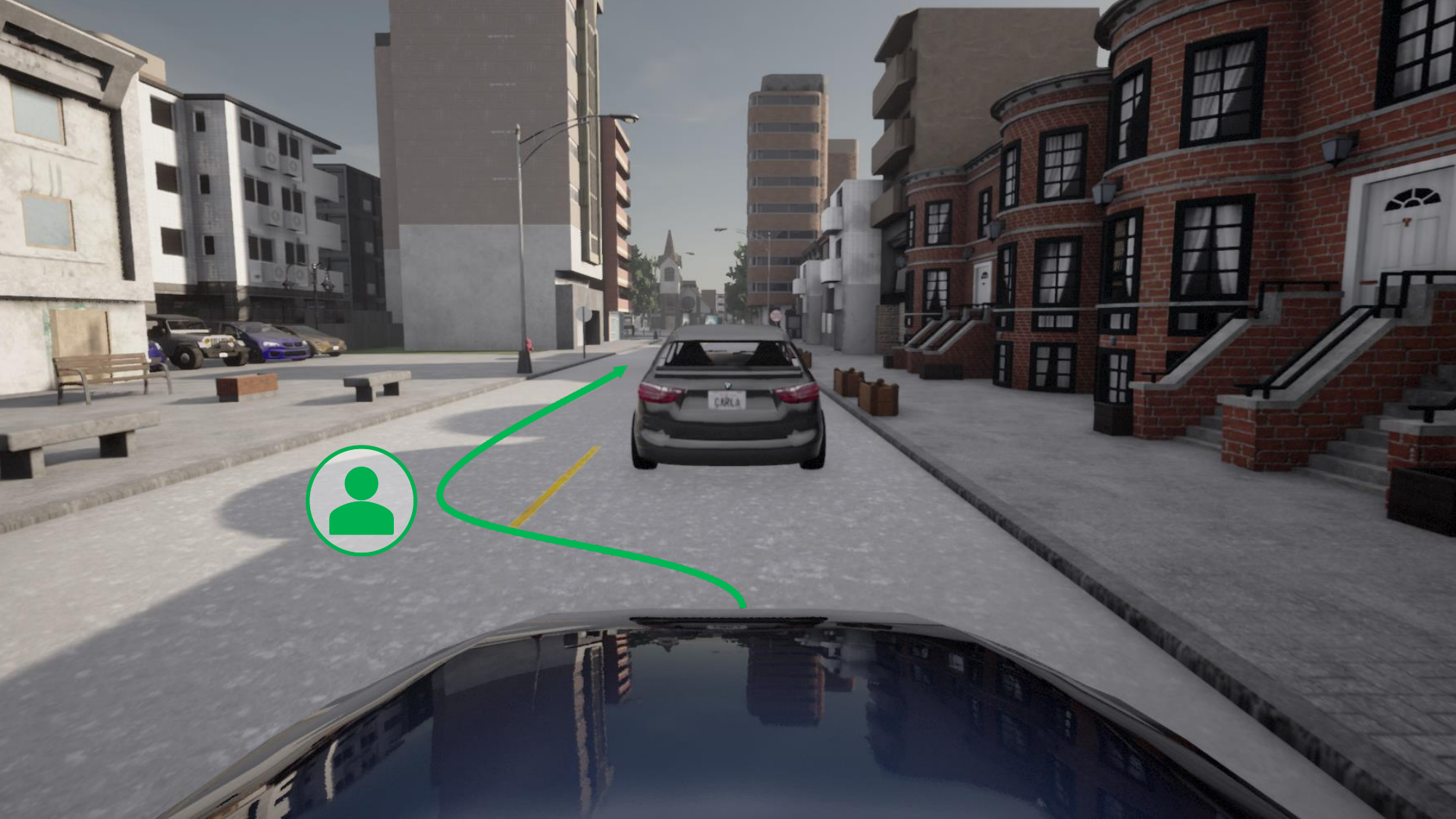}
        \\[-0.6 cm]
        \caption{Direct Control}
        \label{fig:dc}
    \end{subfigure}%
    \hfill
    \begin{subfigure}{0.33\textwidth}
        \includegraphics[width=\linewidth, trim={4.5cm 3.5cm 4.5cm 3.5cm}, clip]{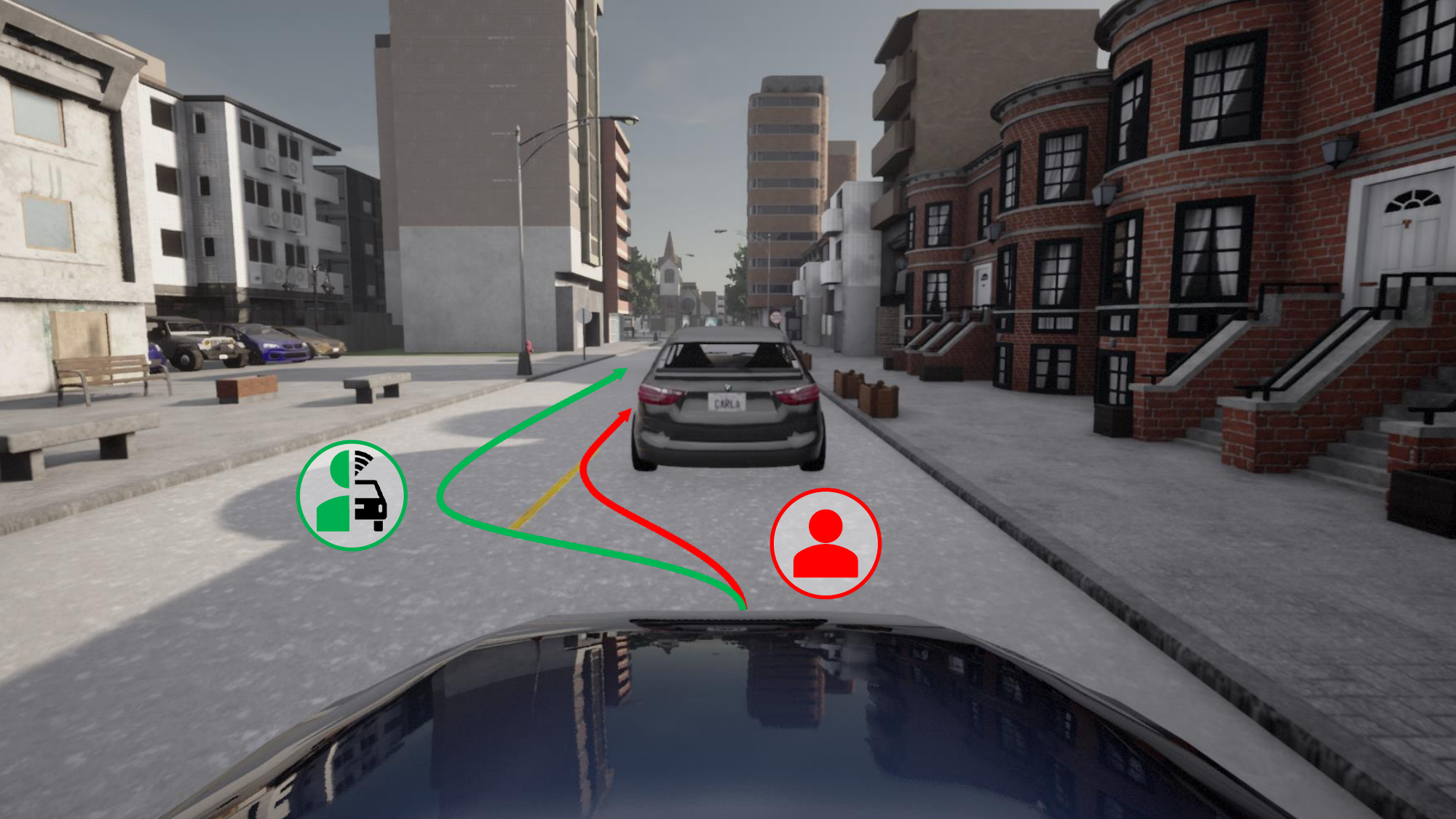}
        \\[-0.6 cm]
        \caption{Shared Control}
        \label{fig:sc}
    \end{subfigure}%
    \hfill
    \begin{subfigure}{0.33\textwidth}
        \includegraphics[width=\linewidth, trim={4.5cm 3.5cm 4.5cm 3.5cm}, clip]{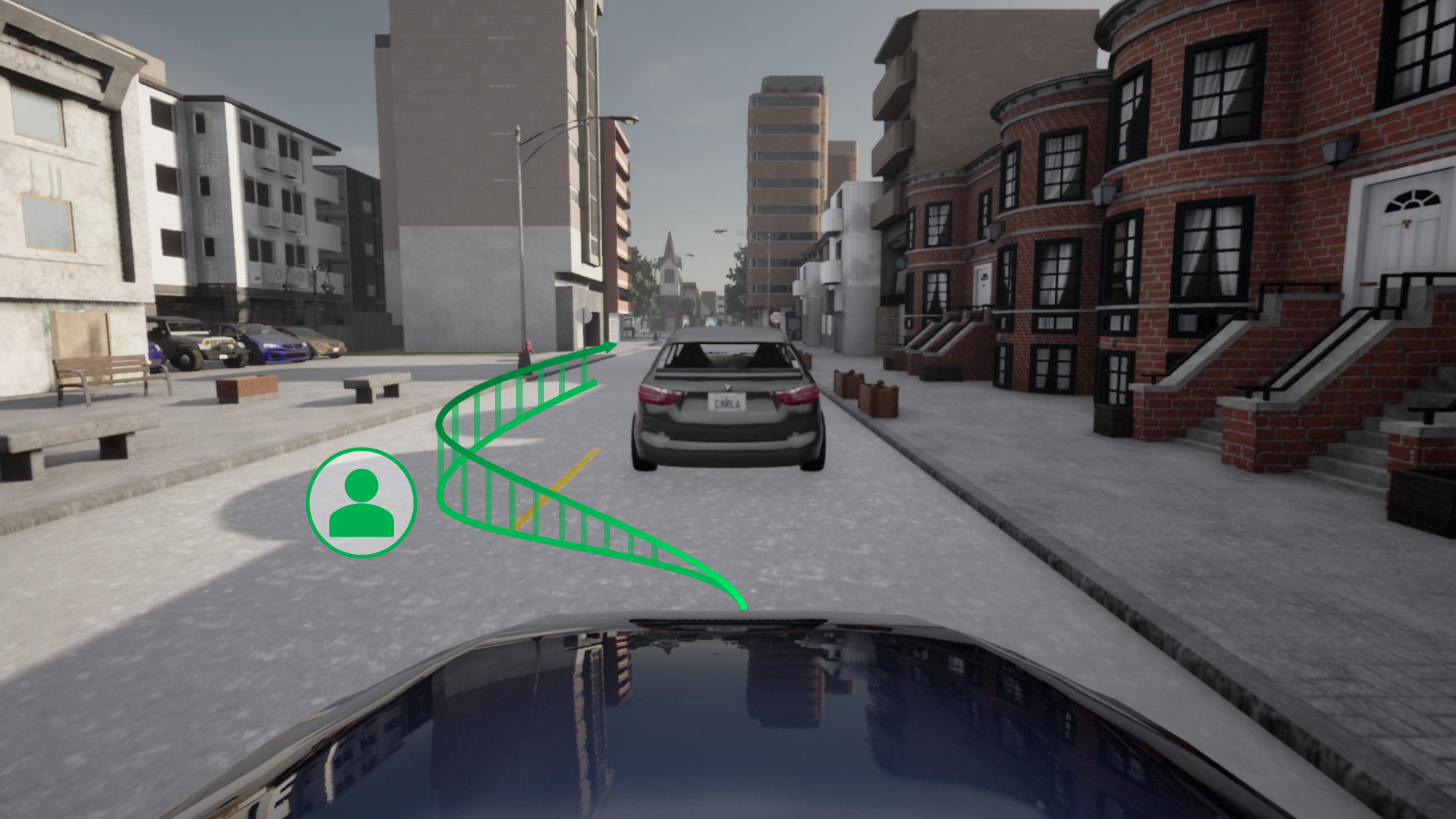}
        \\[-0.6 cm]
        \caption{Trajectory Guidance}
        \label{fig:tg}
    \end{subfigure}
    \\[-0.2 cm]
    \caption{Overview of Remote Driving Teleoperation Concepts}
    \label{fig:test}
\end{figure*}

The \ac{av} disengagements researched in the previous section can be potentially resolved by a human remote operator through teleoperation. Majstorovic \textit{et al.} \cite{Maj2022b} illustrated the extent to which the operator replaces specific \ac{av} modules using the \textit{Sense-Plan-Act} paradigm, representing the general Automated Driving System pipeline as shown in \cref{fig:concepts}. 
This section briefly outlines the \acp{tc} that can support the AV and help solve the disengagement. 
Additionally, an overview of existing studies comparing the introduced \acp{tc} with respect to corresponding \acp{kpi} is presented.

\subsection{Teleoperation Concepts}
Considering the degree of human intervention with respect to control of DDT during teleoperation, Teleoperation Concepts can be grouped into two distinct categories: Remote Driving and Remote Assistance.
The following summarizes the main differences and challenges of the \acp{tc} for each group.

\subsubsection{Remote Driving}
\textit{Remote Driving} groups three \acp{tc} where the DDT is completely under remote control.
Here, the operator usually uses a steering wheel and gas/brake pedals to generate control commands for the vehicle. 

In \textit{Direct Control}, the RO operates the vehicle directly by sending the desired velocity and steering signals based on available environmental data, i.e., video stream(s) and other sensor data (\cref{fig:dc}). While mostly efficient, this concept suffers from reduced operator's situational awareness \cite{Mut2021}, high mental workload, and sensitivity to network issues such as latency or data transmission losses \cite{Geo2020}.

\begin{figure*}[hbp]
    \centering
    \begin{subfigure}{0.33\textwidth}
        \includegraphics[width=\linewidth, trim={4.5cm 3.5cm 4.5cm 3.5cm}, clip]{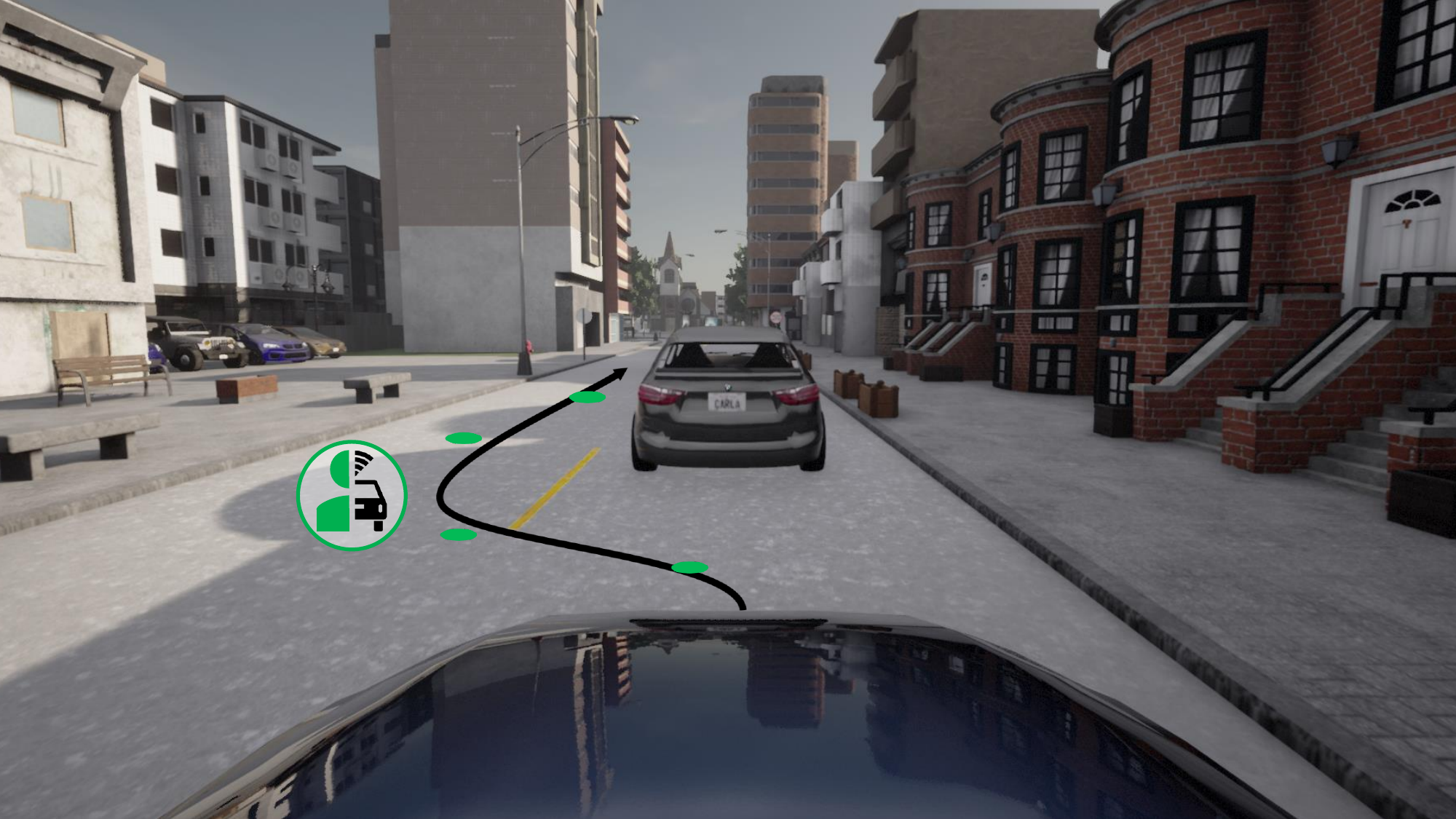}
        \\[-0.6 cm]
        \caption{Waypoint Guidance}
        \label{fig:wg}
    \end{subfigure}%
    \hfill
    \begin{subfigure}{0.33\textwidth}
        \includegraphics[width=\linewidth, trim={4.5cm 3.5cm 4.5cm 3.5cm}, clip]{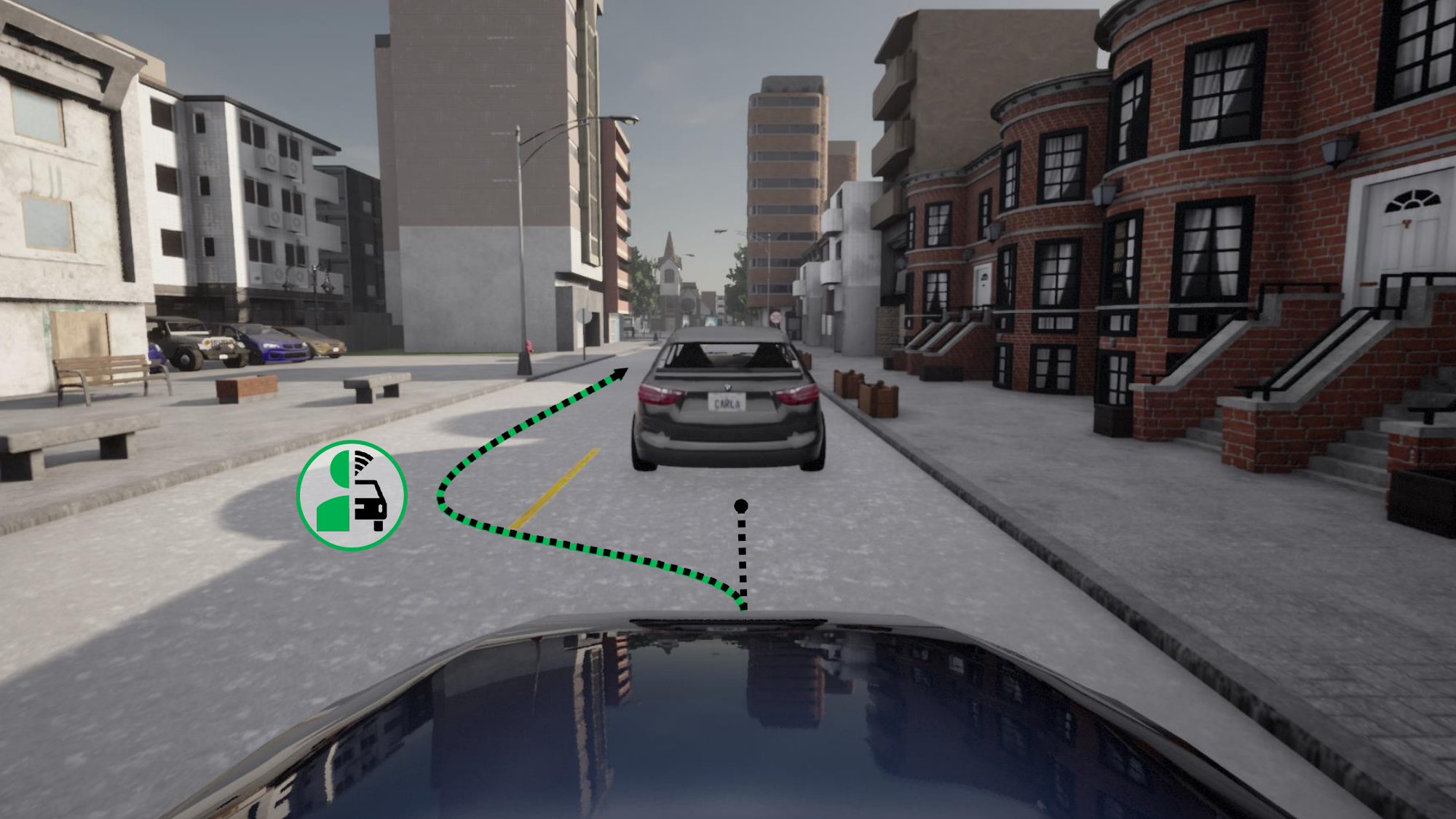}
        \\[-0.6 cm]
        \caption{Collaborative Planning}
        \label{fig:cp}
    \end{subfigure}%
    \hfill
    \begin{subfigure}{0.33\textwidth}
        \includegraphics[width=\linewidth, trim={4.5cm 3.5cm 4.5cm 3.5cm}, clip]{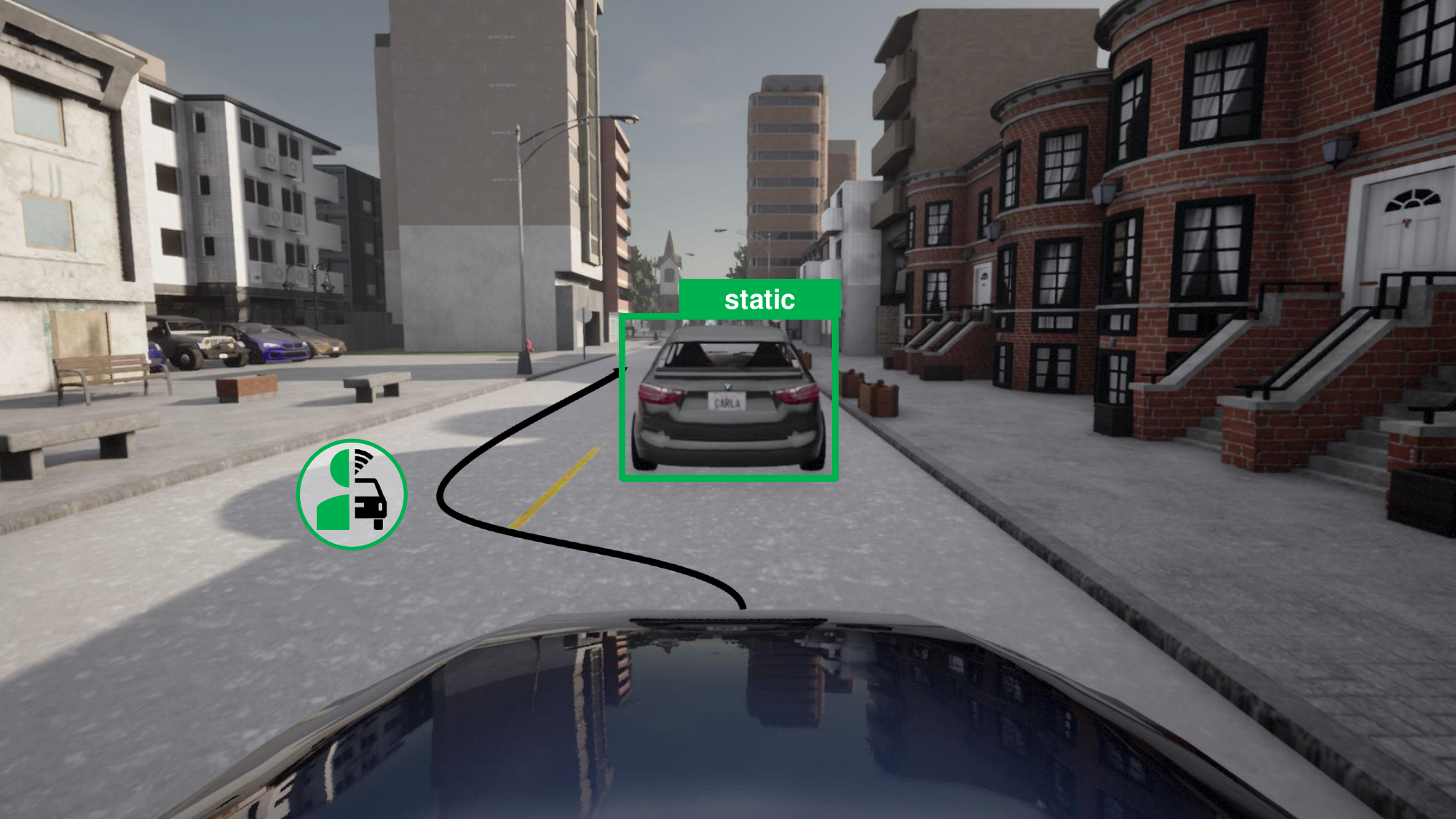}
        \\[-0.6 cm]
        \caption{Perception Modification}
        \label{fig:pm}
    \end{subfigure}
     \\[-0.2 cm]
    \caption{Overview of Remote Assistance Teleoperation Concepts}
    \label{fig:test2}
\end{figure*}

Motivated by these shortcomings, \textit{Shared Control} can be seen as an enhanced version of Direct Control with additional safety measures \cite{Schi2021}.
Here, the operator inputs are evaluated against different safety objectives, e.g., collision avoidance (\cref{fig:sc}). If the operator's input does not fulfill them, the vehicle controller will overrule and adapt it with a set of safe commands to meet the objectives \cite{Sch2022b}. While beneficial from the safety perspective, this mixed control strategy can induce additional workload on the operator to understand the system actions, which puts new requirements on the HMI design for this concept \cite{Yun2023}.

While Shared Control can successfully improve different safety objectives, it is still vulnerable to network issues. To cope with this, Kay \textit{et al.}  \cite{Kay1995} devised an idea to decouple the vehicle stabilization task and let the human operator define trajectories for the vehicle to follow in a \ac{tc} known as \textit{Trajectory Guidance}. The remote interaction is realized in two steps: trajectory creation (remote operator) and trajectory following (\ac{av}) - effectively reducing side-effects of network latencies to a minimum (\cref{fig:tg}). On the negative side, the decoupling can increase overall interaction time, possibly affecting the operator's workload.

\subsubsection{Remote Assistance}
In contrast to Remote Driving, \textit{Remote Assistance} \acp{tc} aim to solve event-driven tasks that are typically executed within a significantly shorter time span.
Here, the primary role of the operator is to advise or guide the \ac{av} system by providing high-level assistance at the perception or planning level while the system retains the driving task.
This significant operator task reduction focuses on increasing the system safety by allowing an operator to make efficient decisions while being relieved of the driving responsibility.

In \textit{Waypoint Guidance}, the operator specifies waypoints the vehicle should follow as demonstrated in the project 5GCroCo \cite{5gcroco2021} or by Schitz \textit{et al.} \cite{Schi2020}. The AV validates the given path and generates an appropriate trajectory for the vehicle to follow (\cref{fig:wg}). 
By assisting the vehicle at the planning level, the human RO can utilize the potential to solve tasks efficiently and with a reduced workload. 
On the negative side, the overall interaction time poses a challenge, often producing unwanted stop-and-go driving behavior.

Motivated by this major shortcoming, the \textit{Collaborative Planning} 
combines information from different perception modalities and formulates path suggestions at the behavior level for the human operator to choose from (\cref{fig:cp}). This level of remote interaction further reduces the operator's responsibilities and therefore offers potential improvements over Waypoint Guidance.
The initial idea has been introduced by Hosseini \textit{et al.} \cite{Hos2014} and improved by Schitz et al. \cite{Schi2021a}.
Majstorovic \textit{et al.} \cite{Maj2023} offered an implementation that supported the full integration within the \ac{av} stack, extending the overall concept with the ODD modification idea to possibly improve the operator's efficiency and situational awareness while reducing the workload. Finally, the term "collaborative planning" was introduced, highlighting
this human-machine synergy to reach a common objective, which we adopt.

To further reduce the responsibility of the operator, the \textit{Perception Modification} concept aims to enable support to the \ac{av} on the perception level. By modifying the vehicle's current environmental model to match the real environment the operator may correct false detections or misinterpretations of the real world. Thus, access to raw sensor data, along with the current environmental model and data from the planning level is required such that the operator can carry out well-founded modifications (\cref{fig:pm}). A first version of the concept was proposed in \cite{Fei2021} and was developed to solve situations in which either a false-positive or an indeterminate and negligible object detection prevents an \ac{av} from continuing its mission. Besides this use-case, the concept can be extended to allow a interaction with any parts of the environmental model of an \ac{av} (e.g. modification of further object properties, the HD~Map, the drive-able space or prediction of object movements) \cite{Fei2023}.

\subsection{Comparison of Teleoperation Concepts} \label{subsec:kpi-survey}

In order to find out which \ac{tc} is most suitable for several \ac{av} disengagements and derive a holistic set, a literature survey is carried out which highlights existing evaluations of the different teleoperation concepts as well as existing comparisons between those concepts. 

An overview of prior evaluations of the concepts is depicted in \cref{tab:evalToD}. Direct Control is the most evaluated concept while Remote Assistance concepts have been evaluated far less than those from the Remote Driving category.

\begin{table}[!h]
    \centering
    \small
    \caption{Evaluations of Teleoperation Concepts}
    \begin{tabular}{l c c}
         \toprule
         Direct Control &   \hspace{-15mm} & \makecell{\cite{App2007}      %
                    \cite{Are2022}      %
                    \cite{Chu2014}      %
                    \cite{Dav2010}      %
                    \cite{Sat2021} %
                    \cite{Zha2020}      %
                    \cite{Fei2020}      %
                    \cite{Geo2018}      %
                    \cite{Nak2021}      %
                    \cite{Tan2014}} \\    %
         Shared Control & \hspace{-15mm} &\makecell{\cite{Sch2020b}     %
                    \cite{Qia2021}      %
                    \cite{Schi2021}     %
                    \cite{Sch2022b}} \\   %
         Trajectory Guidance &  \hspace{-15mm} &\makecell{\cite{Gna2012}      %
                    \cite{Hof2022b}     %
                    \cite{Kay1997}  
                    \cite{Tan2014}      %
                    \cite{Kim2013}      %
                    \cite{Pra2023}} \\     %
         Waypoint Guidance & \hspace{-15mm} &\makecell{\cite{Schi2020}} \\
         Collaborative Planning & \hspace{-15mm} &\makecell{\cite{Hos2014}      %
                    \cite{Schi2020}     %
                    \cite{Maj2023}} \\     %
         Perception Modification &  \hspace{-15mm} &\makecell{\cite{Fei2021}} \\
         \bottomrule
    \end{tabular}
    \label{tab:evalToD}
\end{table}

Prior comparisons between \acp{tc} are depicted in \cref{tab:cmpToD}. It is noticeable that overall, there are significantly fewer studies comparing teleoperation concepts against each other in comparison to the individual and isolated evaluation of the concepts which was shown above. Furthermore, comparisons and studies were only carried out against the Direct Control concept.
\begin{table}[!h]
    \centering
    \small
    \caption{Comparisons of Teleoperation Concepts}
    \begin{tabular}{l c}
         \toprule
         Direct Control $\leftrightarrow$ Shared Control &  \makecell{\cite{And2013}      %
                    \cite{Hos2016c} 
                    \cite{Hos2016b} 
                    \cite{Sto2017}}  \\    %
         Direct Control $\leftrightarrow$ Trajectory Guidance & \makecell{\cite{Pra2023}} \\   %
         Direct Control $\leftrightarrow$ Perception Modification & \makecell{\cite{Fei2023}} \\
         \bottomrule
    \end{tabular}
    \label{tab:cmpToD}
\end{table}
The studies for comparisons of Direct Control and Shared Control show that the number of collisions could be reduced \cite{And2013,Hos2016c} and depending on the implementation Shared Control can also increase efficiency \cite{And2013,Sto2017}. 
Trajectory Guidance is compared to Direct Control in one publication \cite{Pra2023}. This simulative study points out that Trajectory Guidance has higher robustness against external disturbances such as latency or low adhesion but a slower competition time compared to Direct Control. 
Furthermore, there is only one comparison Perception Modification to Direct Control \cite{Fei2023}. The study shows that the Perception Modification concept imposes less mental load on the operators compared to Direct Control.

Prior comparisons and evaluations include research on Direct Control, which is the most evaluated and only \ac{tc} that has been compared to different other concepts.
So far Shared Control has only been compared with Direct Control. 
The studies show that the number of collisions could be reduced \cite{And2013,Hos2016c} and depending on the implementation Shared Control can also increase efficiency \cite{And2013,Sto2017}. 
There is only one publication that evaluates a Trajectory Guidance implementation in comparison to Direct Control \cite{Pra2023}. 
This study reveals that Trajectory Guidance shows higher robustness against external disturbances such as latency or low adhesion but a slower competition time compared to Direct Control.
Also, it must be noted that the experiments are solely conducted in simulation.
Furthermore, there is only one comparison of a Remote Assistance concept \cite{Fei2023}. 
The study carried out compares a implementation of Perception Modification with Direct Control and shows that the Perception Modification concept imposes less mental load on the operators compared to Direct Control.

\cref{table:kpis} lists the collection of \acp{kpi} identified in the literature survey which were used to evaluate \acp{tc}. 
Further \acp{kpi} that evaluate other parts of the teleoperation system, such as network quality or latencies, are omitted 
as they are not applicable to the evaluation of the concepts.
The collected \acp{kpi} were categorised into \textit{Objective} and \textit{Subjective} indicators.
The Objective-class is further divided into the sub-classes \textit{Efficiency}, \textit{Safety} and \textit{Others}.
It should be noted that not all \acp{kpi} are applicable to every \ac{tc}. For example, the Safety \ac{kpi} \textit{Steering wheel reversal rate} is not available for concepts without steering intervention.

\section{\MakeUppercase{Expert Workshop}}
\label{sec:workshop}

In the expert workshop, the gaps considering the comparison of \acp{tc} are filled.
The first part gives a brief overview of the scenarios and the \acp{kpi} being used as well as the participating experts and their background in the field of teleoperation. 
The next part describes the procedure of the workshop, followed by its results.

\begin{table}[htp]
	\caption{KPIs to evaluate teleoperation concepts}
	\label{table:kpis} 
	\centering
    \small
	\begin{tabularx}{0.49 \textwidth}{c@{ }c  X c @{ } P{1.28cm}} %
		\toprule
		  & & \textbf{KPI} & \textbf{Unit} & \multicolumn{1}{c}{\textbf{Ref.}} \\		
		\midrule
		
		\multirow{29}{*}{\rotatebox{90}{\textbf{Objective}}}                 %
            & \multirow{10}{*}{\rotatebox{90}{\textbf{Efficiency}}}           %
                & Course completion time                  & s	    & \cite{Kay1997}, \cite{Kim2013} \cite{And2013}, \cite{Pra2023} \\
                & & Average velocity                      &\si{\metre\per\second}	    & \cite{Dav2010,And2013} \\
                & & Covered distance                      & m	    & \cite{Kay1997} \\
                & & Area covered                          & \%	    & \cite{Sto2017} \\
		     	& & Number of objects of interest found   & -	    & \cite{Sto2017} \\
                & & Task completion time                  & s	    & \cite{Are2022,Fei2023} \\
                & & Acceptance time                       & s	    & \cite{Are2022,Fei2023} \\
                & & Decision time                         & s	    & \cite{Are2022,Fei2023} \\
                & & Execution time                        & s	    & \cite{Are2022,Fei2023} \\
            \cmidrule(lr){2-5}
            & \multirow{11}{*}{\rotatebox{90}{\textbf{Safety}}}           %
                & Number of collisions	                  & -	    & \cite{And2013}, \cite{Hos2016c} \cite{Sto2017} \\	
		        & & Time to collision                     & s       & \cite{Hos2016b} \\
                & & Time integrated time to collision     & \si{\second\squared}   & \cite{Hos2016b} \\
                & & Steering wheel reversal rate          & -       & \cite{Hos2016c} \\
                & & Driver steer volatility               & rad     & \cite{And2013} \\
                & & Steer rate                            & \si{\radian\per\second}   & \cite{And2013,Sat2021} \\
                & & Dependency on stable data connection  & bool    & \cite{Chu2014} \\
                & & Distance to lane boundary after stop  & m       & \cite{Tan2014} \\
                & & Reaction time to secondary task       & s       & \cite{And2013} \\
                & & Operators ability to stop at a line   & m       & \cite{Geo2018} \\
            \cmidrule(lr){2-5}
            & \multirow{8}{*}{\rotatebox{90}{\textbf{Others}}}           %
                & System latency without network latency  & s       & \cite{App2007} \\
                & & Offset to reference trajectory        & m       & \cite{Dav2010}, \cite{Gna2012} \cite{Sat2021}, \cite{Pra2023} \\
                & & Distance to end of path               & m       & \cite{Gna2012} \\
                & & Desired to actual velocity            & \si{\metre\per\second}     & \cite{Hof2022b,Pra2023} \\
                & & Task solving competence               & bool    & \cite{Geo2018,Fei2020} \\
                & & Std. deviation of Operators RR interval     & s       & \cite{Fei2023} \\
                & & Frequency of skin conductance response& \si{\per\second} & \cite{Fei2023} \\
        \midrule
        \multirow{5}{*}{\rotatebox{90}{\textbf{Subjective}}}         %
                & & Cognitive Workload (NASA-TLX)	      & -       & \cite{Hos2016c} \\	
                & & Cognitive Workload (DALI)	          & -       & \cite{Fei2023} \\	
		     	& & Controllability	                      & -       & \cite{And2013} \\
                & & Sense of confidence	                  & -       & \cite{And2013} \\
                & & Percieved maximum drivable velocity   & \si{\metre\per\second}    & \cite{App2007} \\

		\bottomrule
		\end{tabularx} 
\end{table}

\subsection{Selection of Scenarios}
\label{subsec:scenarios}

\begin{figure*}[!hb]
    \centering
    \begin{subfigure}{0.33\textwidth}
        \includegraphics[width=\linewidth, trim={4.5cm 3.5cm 4.5cm 3.5cm}, clip]{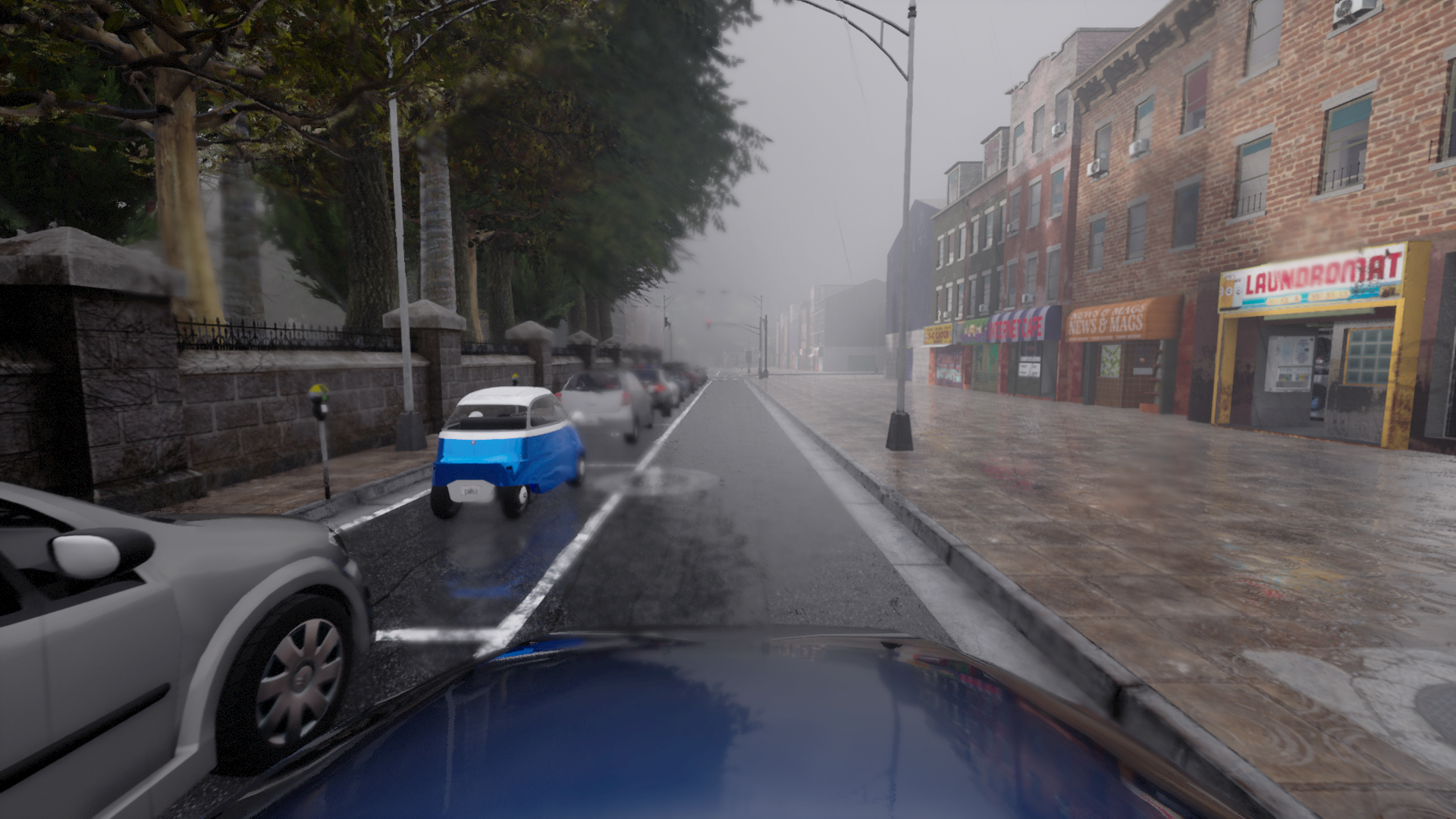}
        \\[-0.6 cm]
        \caption{Scenario 1 -- False-Positive Detection}
        \label{fig:scenario1}
    \end{subfigure}%
    \hfill
    \begin{subfigure}{0.33\textwidth}
        \includegraphics[width=\linewidth, trim={4.5cm 3.5cm 4.5cm 3.5cm}, clip]{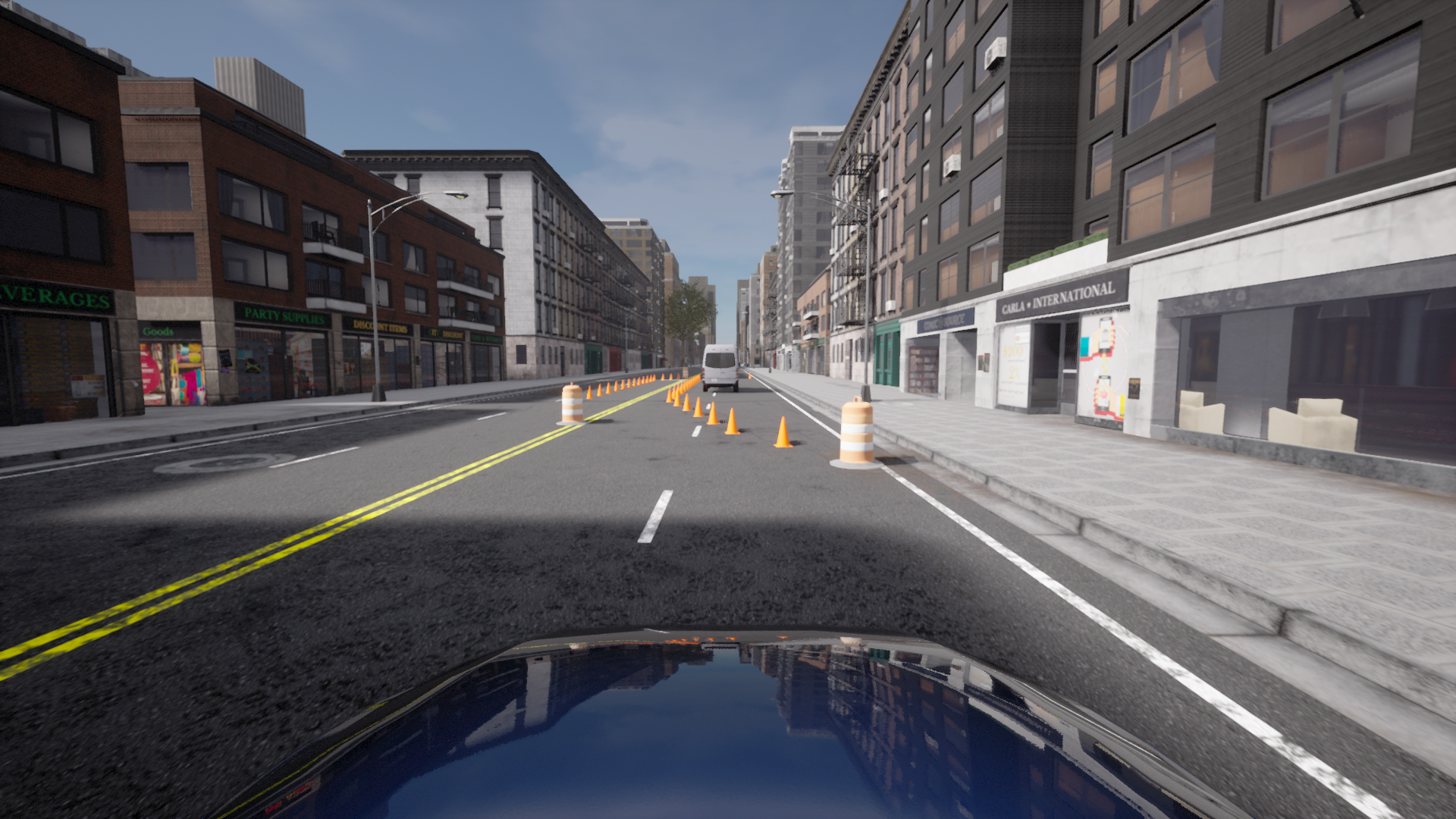}
        \\[-0.6 cm]
        \caption{Scenario 2 -- Construction Site}
        \label{fig:scenario2}
    \end{subfigure}%
    \hfill
    \begin{subfigure}{0.33\textwidth}
        \includegraphics[width=\linewidth, trim={4.5cm 3.5cm 4.5cm 3.5cm}, clip]{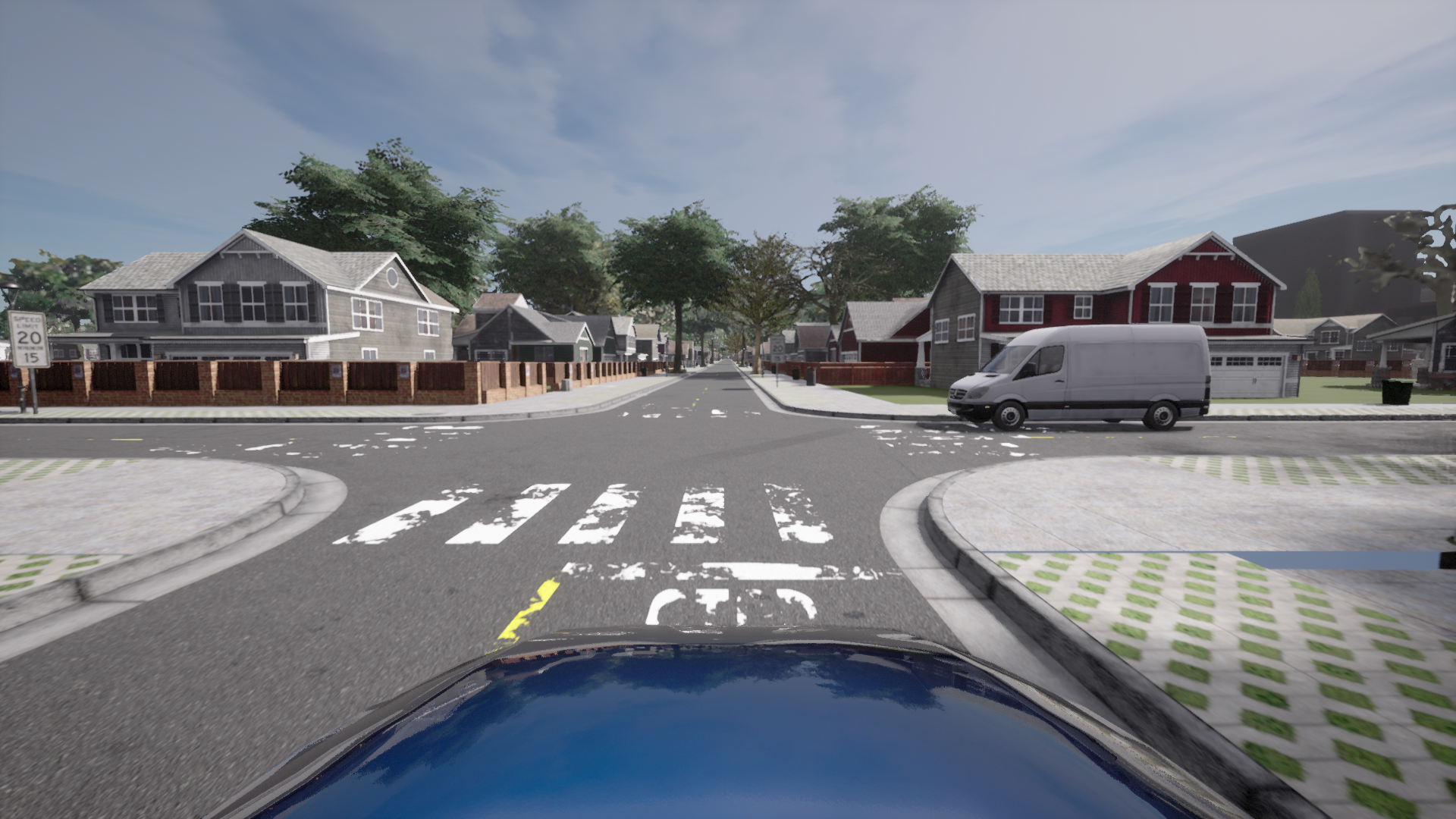}
        \\[-0.6 cm]
        \caption{Scenario 3 -- Second-row parker at intersection}
        \label{fig:scenario3}
    \end{subfigure}
    \\[-0.05 cm]
    \begin{subfigure}{0.33\textwidth}
        \includegraphics[width=\linewidth, trim={4.5cm 3.5cm 4.5cm 3.5cm}, clip]{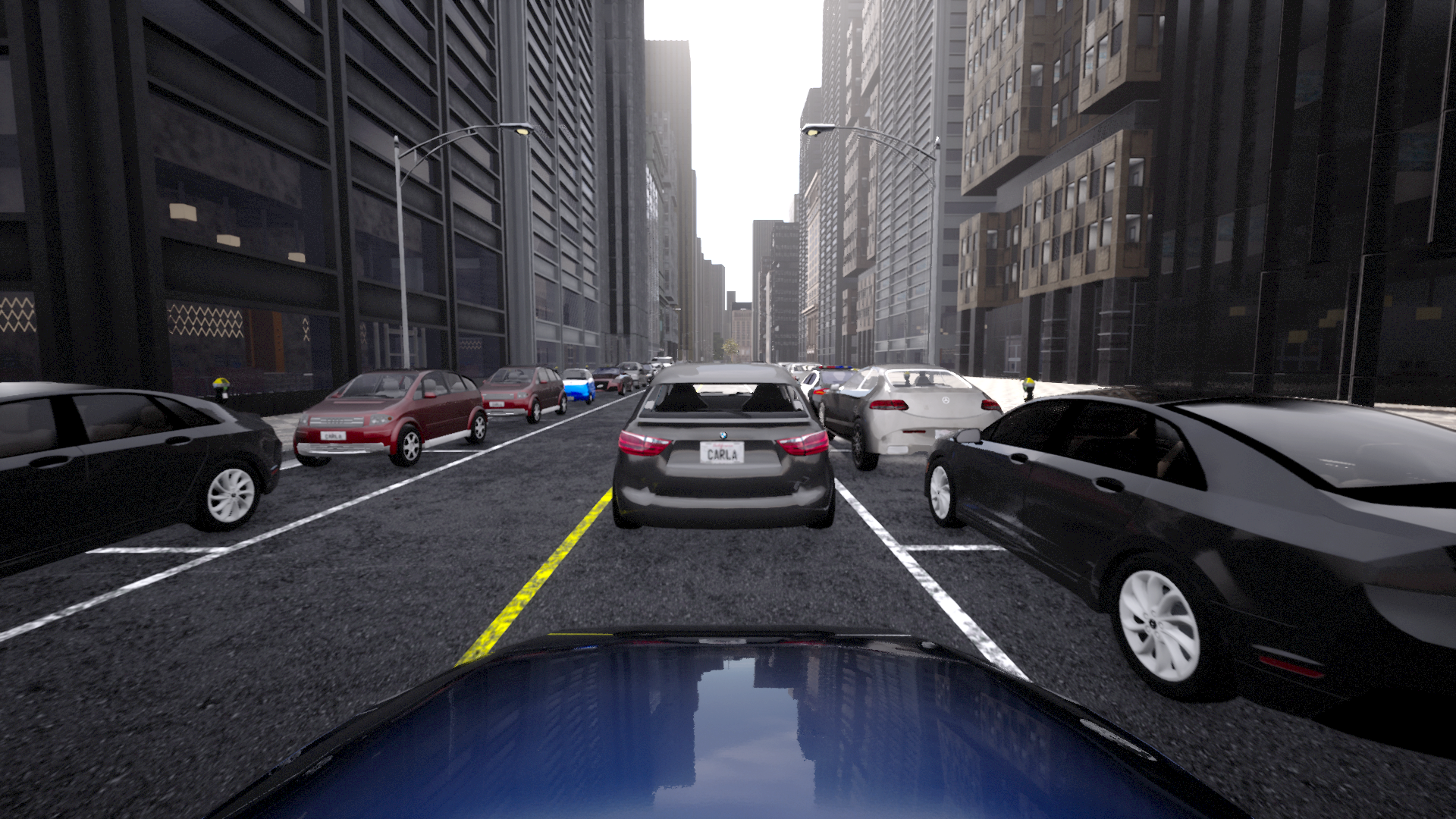}
        \\[-0.6 cm]
        \caption{Scenario 4 -- Lane blocked, solid middle line}
        \label{fig:scenario4}
    \end{subfigure}%
    \hspace{0.00001\textwidth}
    \begin{subfigure}{0.33\textwidth}
        \adjustbox{max height=\height}{\includegraphics[width=\linewidth, trim={4.5cm 3.5cm 4.5cm 4.2cm}, clip]{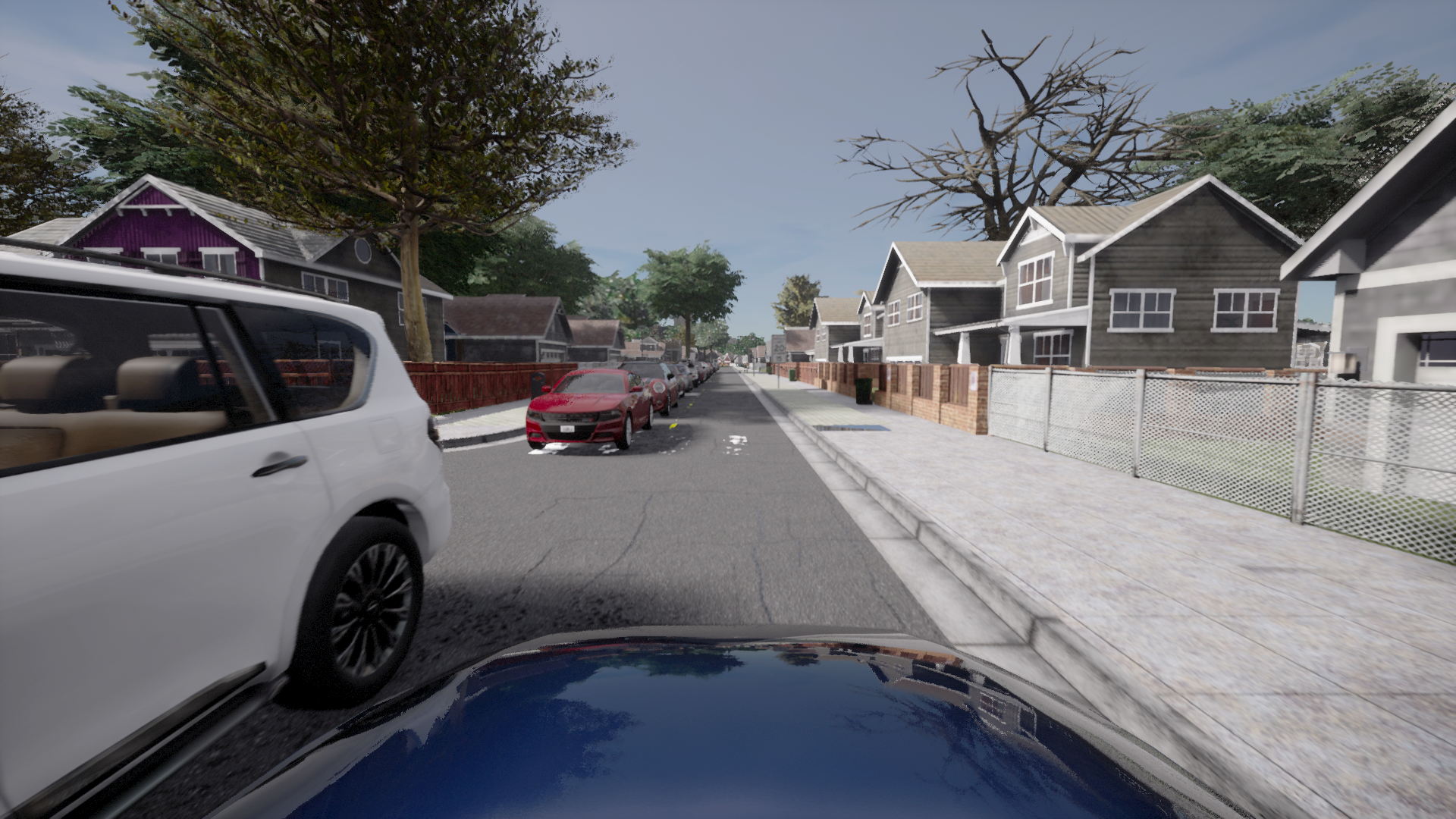}} %
        \\[-0.6 cm]
        \caption{Scenario 5 -- Complex left turn at T-Junction}
        \label{fig:scenario5}
    \end{subfigure}
    \\[-0.2 cm]
    \caption{Overview of disengagement scenarios used in the workshop}
    \label{fig:ScenarioOverview}
\end{figure*}

To compare the \acp{tc} against concrete scenarios, five scenarios were derived from the overview in \cref{sec:avfails}.
The scenarios were selected with the intention to represent a wide range of current \ac{av} disengagements. 
Furthermore, the scenarios were constructed to allow a solution with all \acp{tc}. Each scenario was described textually according to an adapted version of the scenario description template proposed by Schrank~et~al. \cite{schr2021} and reproduced in the CARLA Simulator \cite{Dosovitskiy17}. The derived recordings of the scenarios are in two parts. First, the \ac{av} approaches the unsolvable scene until it comes to a standstill in front of the scene. Part two shows the \ac{av} following the intended solution trajectory. These recordings can be accessed at \url{https://youtu.be/bvjAL_KVEss}.

\textbf{Scenario 1:} This scenario (\cref{fig:scenario1}) involves an failure in the perception module of the Automated Driving System.
A false-positive object is detected ahead of the vehicle in a one-way street. 
This is due to the poor weather conditions and causes the \ac{av} to stop and block the one-way street. 
The operator needs to assist the \ac{av} to continue its path along the on-way street by ignoring the false-positive detection.

\textbf{Scenario 2:} The second scenario (\cref{fig:scenario2}) also represents an failure in the perception module. In this case the \ac{av} is unable to detect lanes. This is due to an unmapped construction site with a changed road layout, resulting in the \ac{av} stopping in front of the construction site. The \ac{av} requires assistance by the operator to continue its ride through the roadworks in accordance with the lanes indicated by the cones.

\textbf{Scenario 3:} Scenario three (\cref{fig:scenario3}) constitutes of a trajectory planning failure in the Automated Driving System. The \ac{av} is unable to find a feasible trajectory to cross an intersection. This is due to an incorrectly parked vehicle at the entry of the intersection, resulting in the \ac{av} getting stuck at the intersection to give the right of way to the parked vehicle. Despite a misinterpretation within the perception module, this scenario represents a trajectory planning error, as it does not enter the junction cautiously after waiting a while, as a human driver would do. The \ac{av} requires assistance by the operator to continue its ride across the junction.

\textbf{Scenario 4:} The fourth scenario (\cref{fig:scenario4}) displays a path planning failure.
The \ac{av} cannot find a drivable path to overtake a parked vehicle. This is due to the other vehicle parking in the second row of a street with a solid center line, resulting in the \ac{av} getting stuck behind the parked vehicle. The operator is required to assist the \ac{av} to bypass the parked vehicle by crossing the solid center line.

\textbf{Scenario 5:} The fifth scenario (\cref{fig:scenario5}) depicts another trajectory planning failure. The \ac{av} cannot find a trajectory to take a left turn at a T-junction. This is due to a busy oncoming lane, resulting in the \ac{av} getting stuck at the T-junction and not driving through the available gap. The operator is required to assist the \ac{av} to turn left through the existing gap.

\subsection{Selection of KPIs}
\label{subsec:kpi-selection}
While \cref{table:kpis} shows the \acp{kpi} used in literature to evaluate \acp{tc}, an appropriate subset 
is chosen for the expert workshop. At least, for each cluster in \cref{table:kpis} one \ac{kpi} is selected taking into account  that these must facilitate a subjective evaluation by the experts and be applicable for all \acp{tc}. In the following, all \acp{kpi} selected for the workshop are described in detail.

\textbf{Efficiency:} The \textit{Task Completion Time} is used to evaluate concepts considering their efficiency. This \ac{kpi} describes the time span between the display of the requested teleoperation on the operator's screen and handing back over the control to the automation of the \ac{av} without the need to monitor it further. This measure includes the acceptance of the teleoperation task by the operator, the situation assessment, the decision on an appropriate action, the execution of the selected action as well as a proper disconnection.

\textbf{Safety:} There are multiple \acp{kpi} when it comes to the safety of road vehicles.
Although being used in the context of vehicle safety, those metrics have shortcomings regarding their comparability. %
These metrics evaluate the safety of a specific scenario.
When comparing different concepts against each other, it can be stated that the final trajectory followed by the supported vehicle in the given scenario does not differentiate amongst different \acp{tc}.
Therefore, the usage of the reviewed \acp{kpi} would lead to equal results for each concept.
This would make it obsolete to evaluate them. Therefore, the authors propose a new \ac{kpi} to overcome those shortcomings, the \textit{Risk induced by Erroneous Human Input}. 
This \ac{kpi} evaluates how much risk is induced to the scenario if the operator makes a mistake.
Possible erroneous human input could be the selection of a wrong GUI-Element or inaccuracies when providing inputs.

\textbf{Others:} One major challenge of \acp{tc} is to handle unstable network connections. Therefore another \ac{kpi} used is the \textit{Robustness against Network Instabilities}. It evaluates the robustness of the mobile data connection in terms of availability (e.g.~ connection loss), data loss (e.g.~ packet loss), latency, and varying bandwidth.

\textbf{Subjective:} The first \ac{kpi} used to evaluate subjective measures of the operator is the \textit{Mental Demand}. This comprised the mental and perceptual activity required by the operator to solve the scenario with the given \ac{tc}.
In addition, the \textit{Controllability} is used as a subjective  \ac{kpi}. This measure indicates how well the operator can assess the criticality of the situation, select and then execute appropriate countermeasures.

\subsection{Participants}
To conduct the expert workshop, participants with extensive experience in the automotive and teleoperation industry or research were invited.
A total of eight experts participated in the workshop.
\cref{tab:experts} gives a brief overview of their expertise in the field.

\begin{table}[!ht]
    \caption{Professional background of participants in the expert workshop}
    \label{tab:experts}
    \centering
    \small
    \begin{tabular}{c p{0.85\columnwidth}}
        \toprule
          & Description \\
        \midrule
        P1 & Research associate in the field of teleoperated road vehicles, academic background in psychology. \\
        P2 & Senior Software Engineering Manager in a company offering teleoperation services, academic background in informatics. \\
        P3 & Human factors research associate in the field of teleoperated road vehicles, academic background in psychology. \\
        P4 & Co-Founder of a teleoperation-startup, academic background in development and design. \\
        P5 & Research associate in the field of automated and teleoperated road vehicles, academic background in mathematics. \\
        P6 & Research associate in a company providing teleoperation solutions, academic background in quality engineering. \\
        P7 & Research associate in the field of automated and teleoperated road vehicles, academic background in mechanical engineering. \\
        P8 & Software Engineer in the field of autonomous driving, academic background in electrical engineering and teleoperation. \\
        \bottomrule
    \end{tabular}
\end{table}

\begin{table*}[!ht]
\centering
\small
\caption{Ranking of Teleoperation Concepts according to KPIs averaged over all scenarios}
\begin{tabular}{clccccc}
    \toprule
    & & \makecell{Task completion\\time}
    & \makecell{Risk induced by \\erroneous human inp.}
    & \makecell{Robustness against\\ network instability}
    & \makecell{Controllability}
    & \makecell{Mental Demand} \\
    \midrule
    \multirow{3}{*}{\rotatebox{90}{\makecell{Remote\\Driving}}}
    & Direct Control          & \cellcolor{6}6 & \cellcolor{6}6 & \cellcolor{6}6 & \cellcolor{1}1 & \cellcolor{6}6 \\ 
    & Shared Control          & \cellcolor{3}3 & \cellcolor{4}4 & \cellcolor{5}5 & \cellcolor{2}2 & \cellcolor{4}4 \\
    & Trajectory Guidance     & \cellcolor{5}5 & \cellcolor{5}5 & \cellcolor{4}4 & \cellcolor{4}4 & \cellcolor{5}5 \\
    \multirow{3}{*}{\rotatebox{90}{\makecell{Remote\\Assist.}}}
    & Waypoint Guidance       & \cellcolor{4}4 & \cellcolor{2}2 & \cellcolor{3}3 & \cellcolor{3}3 & \cellcolor{3}3 \\
    & Collaborative Planning  & \cellcolor{2}2 & \cellcolor{1}1 & \cellcolor{2}2 & \cellcolor{5}5 & \cellcolor{2}2 \\
    & Perception Modification & \cellcolor{1}1 & \cellcolor{3}3 & \cellcolor{1}1 & \cellcolor{6}6 & \cellcolor{1}1 \\
    \bottomrule
\end{tabular}
\label{tab:result-matrix-1}
\end{table*}
\subsection{Workshop Procedure}
The workshop was held at the Institute of Automotive Technology at the Technical University of Munich.
To form a common understanding, the workshop started with an introduction into the taxonomy used in the course of this paper.
Furthermore, to ensure a common understanding of the abilities, limitations, and principles of the \acp{tc}, they were presented in revised in line with the explanations from \cref{sec:comparison}. 
This was followed by an introduction to the \acp{kpi} used in the workshop to clarify their exact definition. 
In the next step, the experts were shown a scenario represented by the textual description and the video before the evaluation of each \ac{tc}.
By resolving all occurring questions a common understanding of each scenario could be created.

After the presentation of each scenario, the participants started the scenario-based evaluation of the \acp{tc}. 
Each expert rated the \acp{tc} based on the specified KPIs for each scenario in a 20-minute period. 
The evaluations were requested in an online questionnaire using a Likert scale reaching from \textit{very low} (one point) to \textit{very high} (five points) or \textit{not applicable}. 
A 10-minute discussion in the plenum summarised each scenario evaluation. 
The discussions and comments of the participants were audio-recorded to ensure an optimal analysis of the feedback.

\subsection{Results}

In the following, the results of the expert workshop are described.
First, an overview comparing the \acp{tc} related to the selected \acp{kpi} is given. 
Furthermore, the ranking of \acp{tc} by scenario as well as the results of the overall ranking are described.

\subsubsection{Ranking of Teleoperation Concepts according to KPIs}

\begin{table*}[!hb]
\centering
\small
\caption{Ranking of Teleoperation Concepts according to Scenarios}
\begin{tabular}{clccccc}
    \toprule
    & & \makecell{Scenario 1}
    & \makecell{Scenario 2}
    & \makecell{Scenario 3}
    & \makecell{Scenario 4} 
    & \makecell{Scenario 5} \\
    \midrule
    \multirow{3}{*}{\rotatebox{90}{\makecell{Remote\\Driving}}}
    & Direct Control          & \cellcolor{4}4 & \cellcolor{4}4 & \cellcolor{4}4 & \cellcolor{5}5 & \cellcolor{2}2 \\ 
    & Shared Control          & \cellcolor{6}6 & \cellcolor{1}1 & \cellcolor{3}3 & \cellcolor{3}3 & \cellcolor{1}1 \\
    & Trajectory Guidance     & \cellcolor{5}5 & \cellcolor{5}5 & \cellcolor{6}6 & \cellcolor{6}6 & \cellcolor{5}5 \\
    \multirow{3}{*}{\rotatebox{90}{\makecell{Remote\\Assist.}}}
    & Waypoint Guidance       & \cellcolor{3}3 & \cellcolor{2}2 & \cellcolor{4}4 & \cellcolor{4}4 & \cellcolor{3}3 \\
    & Collaborative Planning  & \cellcolor{2}2 & \cellcolor{3}3 & \cellcolor{2}2 & \cellcolor{1}1 & \cellcolor{4}4 \\
    & Perception Modification & \cellcolor{1}1 & \cellcolor{6}6 & \cellcolor{1}1 & \cellcolor{1}1 & \cellcolor{6}6 \\
    \bottomrule
\end{tabular}
\label{tab:result-matrix-2}
\end{table*}
The results of the ranking are depicted in \cref{tab:result-matrix-1} averaged over all scenarios.
As can be seen from the table, remote assistance concepts are generally ranked better than remote driving concepts for all KPIs except controllability. Furthermore, Perception modification is ranked best in three of five KPIs, whereas Direct Control is ranked worst in four of five KPIs. Among all Remote Assistance concepts, Waypoint Guidance is generally ranked worst. 

\textit{Task completion time} was ranked lowest for Perception Modification and Collaborative Planning as these concepts enable a fast and simple interaction with the AV especially if the goal position of the vehicle is clear and in sight of the RO.
The experts assumed that using these TCs, the RO can directly disconnect as soon as the input is given for those scenarios.   

Reasons for \textit{risk induced by human error} being low on remote assistance TCs and especially collaborative planning is the lower involvement of the operator according to the experts. They argued that a low involvement also reduced the risk induced by human error. Shared Control is ranked best across remote driving concepts since it introduces collision avoidance.

According to the experts, the reason for higher \textit{robustness against network instability} in Remote Assistance concepts stems from the type of input given to the vehicle. Remote Assistance uses discrete inputs, whereas Remote Driving uses continuous inputs, which are more vulnerable to network instabilities.

Due to the direct influence the operator has on the actuators of the vehicle in Direct Control and the familiarity of the input methods used in this TC compared to driving in a vehicle, \textit{controllability} was ranked higher for remote driving TCs. One factor that decreases the controllability of Remote Assistance concepts is the general inability to change the decision dynamically after the input is sent to the vehicle.

The \textit{Mental Demand} was ranked lower for Remote Assistance concepts especially is false positive object detections and path planning failures occur.  
Furthermore, the absence of collision avoidance leads to a high mental demand for Direct Control and Trajectory Guidance.
However, if planning the future vehicle path requires high effort through occlusions or dynamically changing situations, Remote Driving becomes more favorable.

\subsubsection{Ranking of Teleoperation Concepts according to Scenarios}
\label{subsec: rankingScenarios}

At the end of each voting round, the experts were asked to rank the \acp{tc} for the specific scenario.  
As can be seen from the results in \cref{tab:result-matrix-2}, there is a clear division between scenarios where \acp{tc} from the Remote Assistance category are ranked higher and scenarios where \acp{tc} from the Remote Driving category are ranked higher.  

For solving \textit{Scenario 1}, \textit{Scenario 3} and \textit{Scenario 4}, Remote Assistance concepts are preferred by the experts. 
It is noticeable that Perception Modification takes first place in all of them and Collaborative Planning takes second place except for \textit{Scenario 4}, where it shares first place with Perception Modification. 
Waypoint guidance is ranked noticeably ranked worse.

\textit{Scenario 2} was rated approximately equally between Remote Driving and Remote Assistance concepts. 
It is noticeable that, in contrast to the previously mentioned scenarios, Shared Control is ranked first while Perception Modification was ranked last with one of the study participants stating that this concept could not solve the scenario at all. 

To solve \textit{Scenario 5}, Remote Driving concepts were clearly favored.
As before, Perception Modification takes last place in this situation, with half of the experts voting that this concept could not solve the scenario at all. 
As stated by the experts, the reasoning for this is that it is not clear what part of perception can be modified in order to solve the disengagement of the \ac{av}.

Overall, it is noticeable that Trajectory Guidance takes fifth or last place in all scenarios and that Shared Control is mostly ranked better than Direct Control.
This evaluation is independent of whether the scenario favors Remote Driving or Remote Assistance. 
According to the experts, this is the case because it was assumed that Direct Control and Trajectory Guidance do not include collision avoidance.
Further, it was also pointed out that Trajectory Guidance has major disadvantages compared to the other Remote Driving concepts such as disability to carry out dynamic maneuvers or changing the planned trajectory while being carried out.
Also, the applicability and potential of Perception Modification strongly depend on the scenario.
In some scenarios, this \ac{tc} came first, in others it came last or was rated as not applicable.

\subsubsection{Overall Ranking}
\begin{figure}[ht]
    \centering
    \hspace*{-0.18cm}
    \begin{tikzpicture}
    \begin{axis}[
        width=1.08\linewidth,
        height=4cm, 
        ybar, 
        ymin=0,
        ymax=6, 
        ylabel={Average Ranking},
        ylabel style={yshift=-17pt, align=center, font=\small},
        xtick=\empty,
        xtick={1,2,3,4,5,6},
        xticklabel style={text width=1.3cm, align=center, font=\scriptsize},
        xticklabels={Shared Control, Collaborative Planning, Perception Modification, Direct Control, Waypoint Guidance, Trajectory Guidance},
        xmin = 0.7,
        xmax = 6.3,
        axis lines=left
    ]
    \addplot[draw=1, fill=1] coordinates {(1, 1.625) (2, 3.0) (3, 3.0) (4, 4.0) (5, 4.0) (6, 5.375)};
    \end{axis}
    \end{tikzpicture}
    \caption{Average overall ranking of teleoperation concepts (lower is better)}
    \label{fig:overallRanking}
\end{figure}
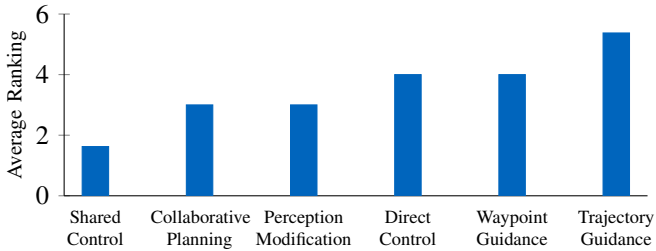

In a sperate overall ranking at the end of the workshop, the experts were asked to give the \acp{tc} a overall ranking after seeing and rating all scenarios. 
Shared Control was first before Perception Modification and Collaborative Planning who both share second place as can be seen from \cref{fig:overallRanking}.
Trajectory Guidance is ranked last, which confirms the trend from the results mentioned in the previous sections. 

In general, it is noticeable that Shared Control takes first place in the overall ranking, although Remote Assistance concepts are favored in four of five \ac{kpi} and in three of five scenarios. 
According to the experts, the reason for that result is the high applicability of Shared Control to different \ac{av} disengagements.

\section{\MakeUppercase{A holistic Set of Teleoperation Concepts}} %
\label{sec:concept}

In the following, a holistic set of \acp{tc} is proposed that is applicable to solve a wide variety of \ac{av} disengagements. 
It is based on the results of the expert workshop. %
Furthermore, limitations arising from the expert workshop's assumptions and format are discussed.

\subsection{Derivation of a Set of Teleoperation Concepts}

The \acp{tc} that form a holistic set of \acp{tc} and are therefore suggested to be implemented are \textit{Shared Control} as well as \textit{Collaborative Planning} and \textit{Perception Modification}. 
The choice of these concepts is explained in the following.

Based on the results from the expert workshop, representatives of both Remote Driving as well as Remote Assistance concepts are needed to enable safe and efficient teleoperation of \acp{av} and solve the widest possible range of disengagements. 
\acp{tc} from both categories perform particularly well in different scenarios and complement each other as described in \Cref{sec:workshop}. 
Based on the characteristics of the presented scenarios as well as the feedback from the experts, it can be concluded that Remote Driving performs better in scenarios  
\begin{itemize}
    \item[(1)] where the problem changes dynamically, e.g., where the \ac{av} is required to pass oncoming traffic while being remotely supported,
    \item[(2)] that include negotiation of a solution with other road participants, or
    \item[(3)] where Remote Assistance concepts would require multiple decision-making steps as well as planning effort, e.g., through low visibility, occlusions, or longer distances over which the vehicle shall be teleoperated.
\end{itemize}

Remote Assistance, on the other hand, performs well in situations
\begin{itemize}
    \item[(1)] where the road segment containing the problem is easy to see and the problem area is clearly defined, visible, and static, or
    \item[(2)] where specific parts of the Automated Driving stack can be supported or replaced by the operator such as the perception or path planning module. 
\end{itemize}

Based on these findings, the implementation of Shared Control is suggested since it addresses \ac{av} disengagements where Remote Driving is preferable and is highly applicable to different scenarios. 
This \ac{tc} is chosen over Direct Control since it was ranked higher in most \acp{kpi} and scenarios. 
Trajectory Guidance is generally not recommended to be included since it was ranked generally much worse than Shared Control and Direct Control and it does not offer any considerable advantages but major disadvantages as described in \cref{sec:workshop}

To address cases where Remote Assistance is preferable, Perception Modification should be implemented as it allows situations such as false-positive object detections to be solved efficiently. 
As shown in \cref{sec:workshop}, Perception Modification is preferred in most scenarios and rated best in the analyzed KPIs, yet it may not be applicable to solve certain disengagements.
Therefore, Collaborative Planning should be implemented since it complements Perception Modification by enabling further \ac{av} disengagements to be addressed.   
The selection of two Remote Assistance Concepts is also related to their working principle. Where Collaborative Planning is related to vehicle guidance through the negotiation of a desired path with the operator, the Perception Modification concept has strengths in assisting other parts of the Automated Driving System such as object and drivable space detection and therefore acts on the logical layer. 
Waypoint Guidance is not considered in the holistic set of \acp{tc} since it addresses similar \ac{av} disengagements as Collaborative Planning but performs worse in all \acp{kpi} except controllability. 

If Collaborative Planning and Perception Modification do not offer a solution in a specific \ac{av} disengagement or any of the conditions for the usage of Remote Driving mentioned above are fulfilled, Shared Control is suggested to be used since according to the expert workshop it is the most widely applicable to solve various fail cases of all \acp{tc} recommended here.

However, there are some challenges when implementing the mentioned concepts in a holistic system. 
Shared Control might not perform well if false-positive object detections are present.
Therefore, it needs to be implemented in a way that enables the operator to correct or overrule the Automated Driving System's perception while performing the driving task. 
Considering the implementation of Collaborative Planning according to the participants in the expert workshop, the handling of false-positive object detections is also an issue that has to be addressed. 
When implementing Perception Modification, it needs to be further investigated what parts of the \ac{av}'s environmental model are meaningful to be modified in order to solve the largest possible set of disengagement causes related to the perception system. 
Since the possibilities for modifying the perception are diverse, e.g., modification of the perceived objects or map data, an optimal set of perception modification modes needs to be determined. 
It is also important to investigate how long the modification should be valid.

\subsection{Limitations}

Due to the assumptions being made and the format of the workshop, there are some limitations to the results that are discussed in the following.

The assumptions include all Automated Driving System modules being functional which might not be fulfilled in certain disengagements and would therefore exclude some \acp{tc} for specific scenarios.
Also, a modular Automated Driving stack comparable to \cref{fig:concepts} was assumed.
As research for end-to-end Automated Driving stacks becomes more popular \cite{chib2023}, further concepts might be needed to solve occurring \ac{av} disengagements for these systems. 
Since there are currently no \acp{tc} addressing such system architectures, they were not part of the present workshop.
Another question to be addressed is whether the operator has to monitor the vehicle while the maneuver is carried out during Remote Assistance since this is an influencing factor on the rating of \acp{kpi}. 

The main limitations of the workshop format arise from the number of participants, the number of scenarios and \acp{kpi} being investigated as well as the conceptual presentation of \acp{tc}. 
To gain further insights, the number of participants should be increased in future evaluations.
Also, the collective of participants should be extended by fleet managers from \ac{av} fleet operations which would give valuable insights into operations of \acp{av} and an additional viewpoint on the implementation of concepts.
Due to the format of the workshop, the number of scenarios was limited to five.
To obtain more reliable results, more scenarios should be compared against.
Also, a comprehensive scenario catalog containing the most common \ac{av} disengagements should be derived. 
To the best of the authors' knowledge, such a catalog does not yet exist. 
In the workshop, the scenarios were only shown in videos from the simulation and the concepts were not implemented in a real-world teleoperation system.
Their functionality was only shown schematically with the goal of not biasing the participants. 
Therefore, expectations across the participants regarding the functionality and implementation of the concepts might vary. 
In the future, all six \acp{tc} examined here should be implemented along a safety concept and user interface in a real-world system to test them against each other with the goal to examine the results from the present paper again. 
A real-world study would also allow to measure further \acp{kpi}.

Despite these limitations, the assessment in this paper shows a clear trend and enables to derive hypotheses and directions for further research.

\section{\MakeUppercase{Conclusion}} %
\label{sec:conclusion}

The scope of the presented work was to derive a set of \acp{tc} to remotely support \acp{av} and to solve a wide range of \ac{av} disengagements.
The set was derived based on a literature analysis on \ac{av} disengagements as well as existing studies on \ac{tc} comparisons and \acp{kpi} to evaluate teleoperation performance.
An expert workshop was conducted in which eight participants from academia and industry rated all \acp{tc} according to a range of scenarios and \acp{kpi}. 
Based on the results, the authors propose that a holistic teleoperation system should be composed of implementations of the Shared Control, Collaborative Planning and Perception Modification concepts.
This set will be needed to solve most of \ac{av} disengagements that occur during operations in a safe and efficient manner.

Future work can consist of the implementation of the different \acp{tc} along user interfaces and safety concepts as part of a control room such that more extensive studies on the concepts and the system's performance can be conducted. The system and its concepts should also be evaluated against quantitative and comparative \acp{kpi} against a larger amount of \ac{av} disengagements.

\section*{\MakeUppercase{Acknowledgements}}
David Brecht, Nils Gehrke, Tobias Kerbl, Niklas Krauss, Domagoj Majstorovi\'c, Florian Pfab, and Maria-Magdalena Wolf, as the first authors, collectively contributed to this work. 
Frank Diermeyer made essential contributions to the conception of the research projects on road vehicle teleoperation and revised the paper critically for important intellectual content. 
He gave final approval for the version to be published and agrees to all aspects of the work. 
As a guarantor, he accepts responsibility for the overall integrity of the paper. 

\begin{acronym}

    \acro{tc}[TC]{Teleoperation Concept}
    \acroplural{tc}[TCs]{Teleoperation Concepts}
    
    \acro{av}[AV]{Automated Vehicle}
    \acroplural{av}[AVs]{Automated Vehicles}
    
    \acro{odd}[ODD]{Operational Design Domain}
    \acro{srtp}[SRTP]{Successive Reference Pose Tracking}
    \acro{kpi}[KPI]{Key-Performance-Indicator}
    \acro{ttc}[TTC]{Time to Collision}       
    \acro{sae}[SAE]{Society of Automotive Engineers}

\end{acronym}

\bibliographystyle{IEEEtran}
\bibliography{IEEEabrv,bibliography}
\vfill
\begin{IEEEbiography}[{\includegraphics[width=1in,height=1.25in,clip,keepaspectratio]{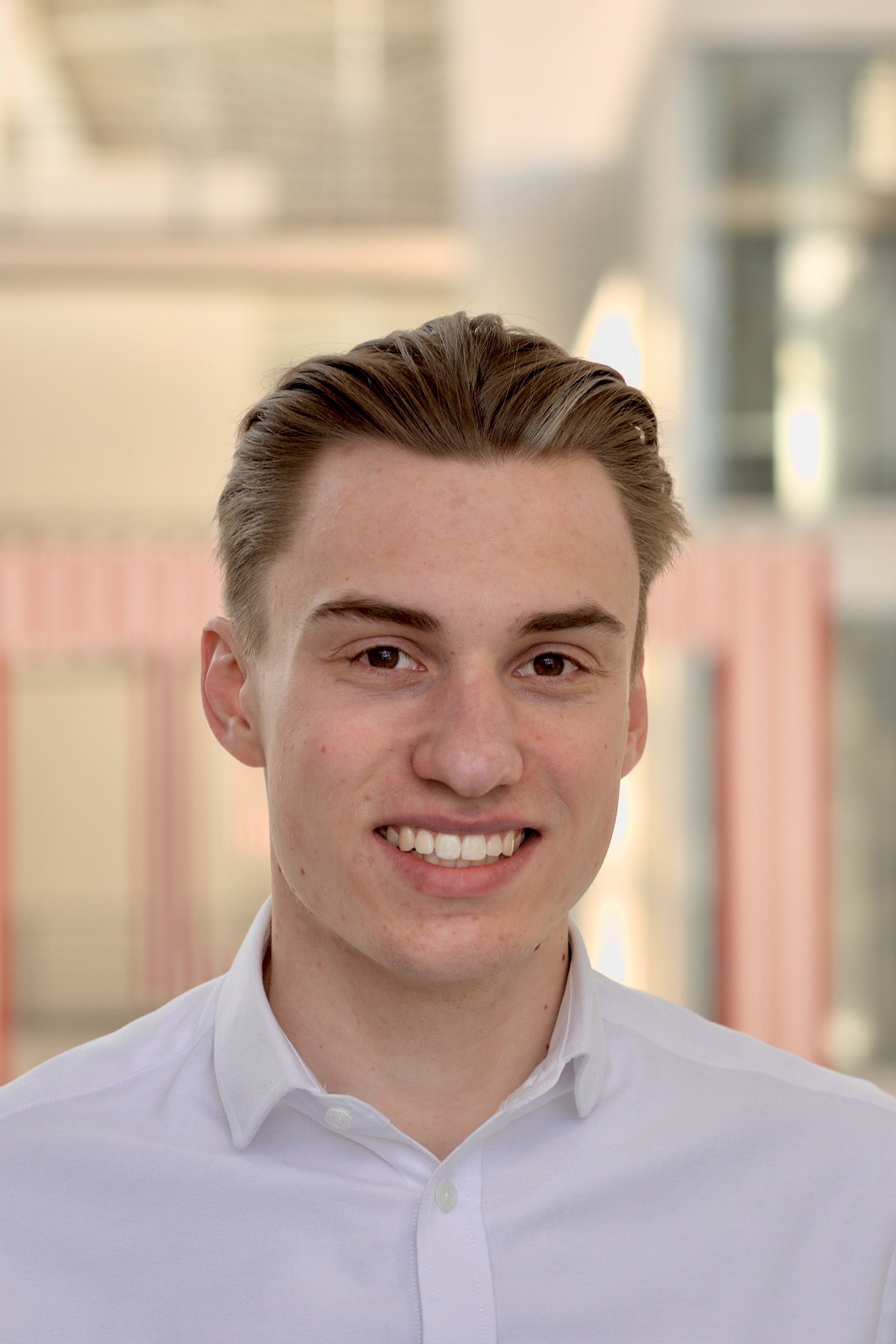}}]{David Brecht } 
received the B.Sc. degree in Mechanical Engineering from Technische Universit\"at Braunschweig in 2020 
and M.Sc. degrees from  Technische Universit\"at Braunschweig and Tongji University in Automotive Engineering in 2022 and 2023 respectively.
He is currently pursuing the Ph.D. degree in Mechanical Engineering with the Institute of Automotive Technology, Technical University of Munich.
His research interests include safety of teleoperated and automated vehicles.
\end{IEEEbiography}

\begin{IEEEbiography}[{\includegraphics[width=1in,height=1.25in,clip,keepaspectratio]{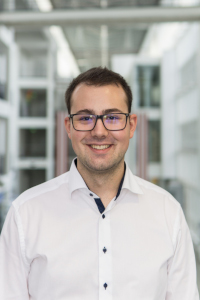}}]{Nils Gehrke } 
holds a B.Sc. degree in Mechanical Engineering from the ETH Zurich and a M.Sc. degree in Computational Science and Engineering from the Technical University of Munich. He is currently pursuing a Ph.D. as a Scientific Researcher at the Institute of Automotive Engineering, where his research focuses on increasing online safety in the communication for remote operated driving using self awareness methods and game theory.
\end{IEEEbiography}

\begin{IEEEbiography}[{\includegraphics[width=1in,height=1.25in,clip,keepaspectratio]{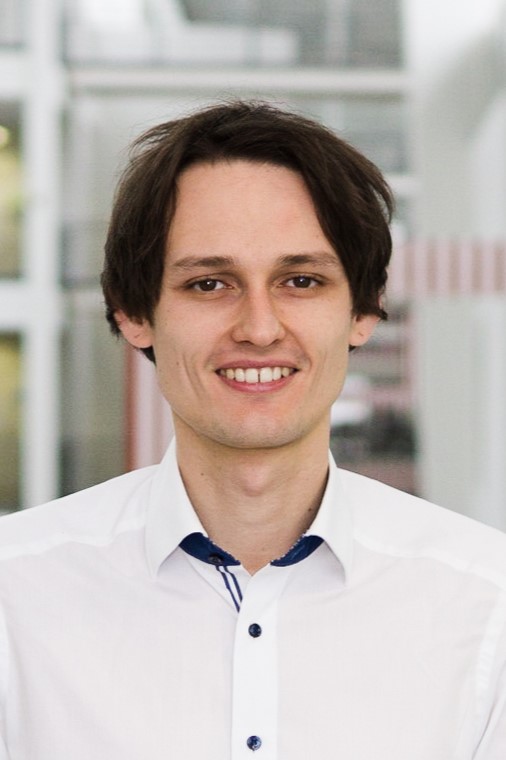}}]{Tobias Kerbl } holds a B.Sc. degree in Mechanical Engineering and a M.Sc. degree in Automotive Engineering from the Technical University of Munich. Currently pursuing a Ph.D. degree in mechanical engineering with the Institute of Automotive Engineering at the Technical University of Munich, his research focuses on the fields of automated vehicles and remote-assistance concepts.
\end{IEEEbiography}

\begin{IEEEbiography}[{\includegraphics[width=1in,height=1.25in,clip,keepaspectratio]{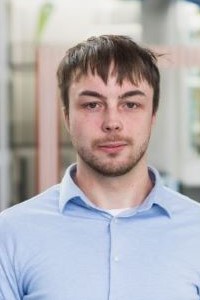}}]{Niklas Krauss } holds a B.Sc. degree in Business Information Systems from the University of Applied Science in Hanover and an international M.Sc. in Data Analytics from the University of Hildesheim. Currently pursuing a Ph.D. at the Institute of Automotive Engineering, his research focuses on the fields of spatial awareness during teleoperation. 
\end{IEEEbiography}

\begin{IEEEbiography}[{\includegraphics[width=1in,height=1.25in,trim=5mm 0 5mm 0, clip,keepaspectratio]{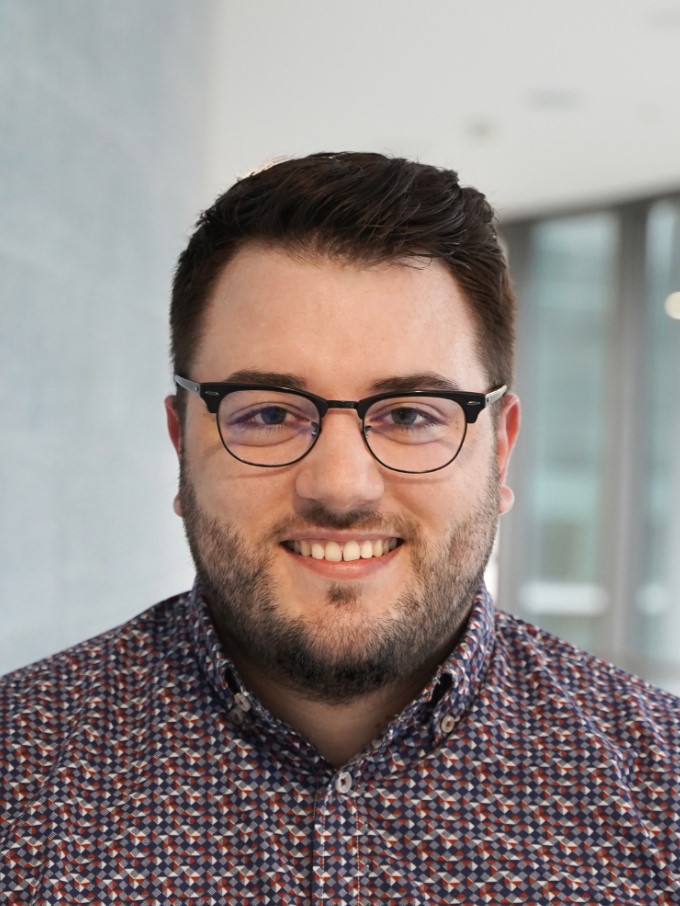}}]{Domagoj Majstorovic} ~holds a M.Sc. degree in Mechanical Engineering from the University of Zagreb, Croatia. Since 2019, he has been pursuing a Ph.D. degree in mechanical engineering with the Institute of Automotive Technology at the Technical University of Munich. His research interests include developing novel assistance systems that help autonomous vehicles cope with challenging driving situations.
\end{IEEEbiography}

\begin{IEEEbiography}[{\includegraphics[width=1in,height=1.25in,clip,keepaspectratio]{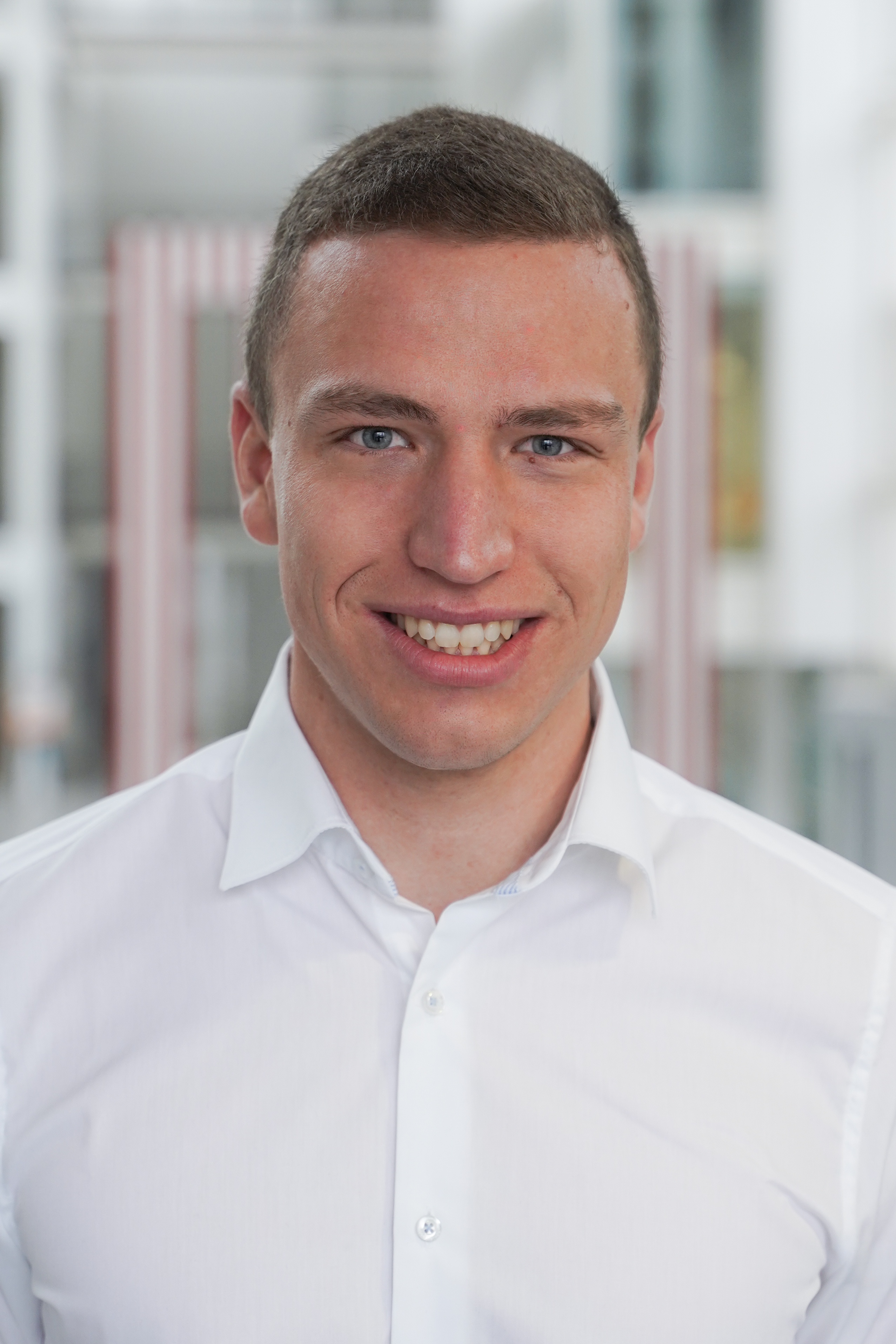}}]{Florian Pfab } 
holds a B.Sc. degree in Mechanical Engineering and a M.Sc. degree in Automotive Engineering from the Technical University of Munich. Currently pursuing a Ph.D. at the Institute of Automotive Engineering, his research focuses on the self-awareness of teleoperated and fully-automated vehicles, with a concurrent role as project lead for the EDGAR-Project.
\end{IEEEbiography}

\begin{IEEEbiography}[{\includegraphics[width=1in,height=1.25in,clip,keepaspectratio]{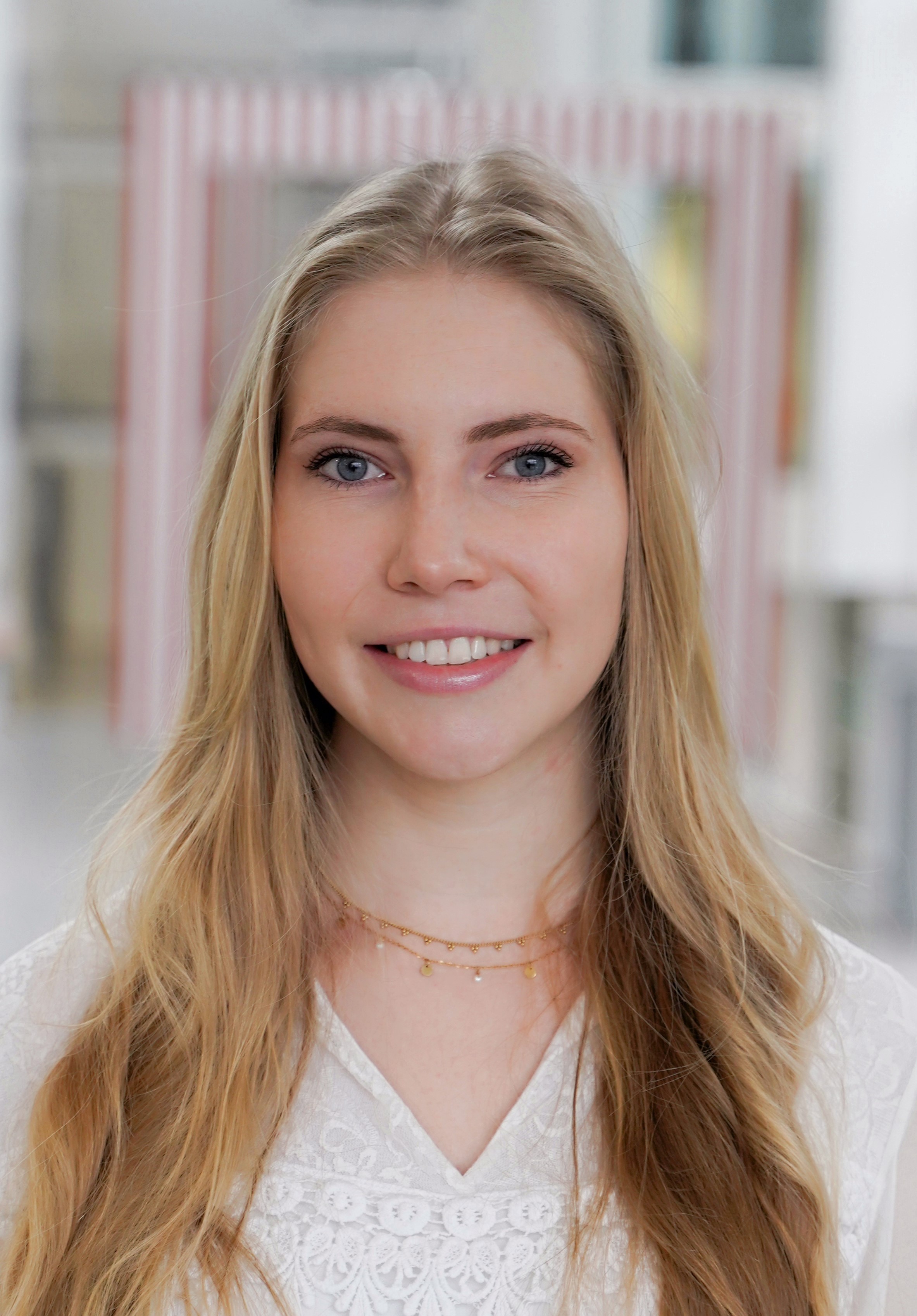}}]{Maria-Magdalena Wolf } received her B.Sc. in Mechanical Engineering and completed her M.Sc. in Automotive Engineering at the Technical University of Munich specializing in human factors and ergonomics. In her doctorate at the Institute of Automotive Technology, Technical University of Munich, she is working on human-machine interaction in teleoperation to optimise the user-friendliness of the human operator's tasks.
\end{IEEEbiography}

\begin{IEEEbiography}[{\includegraphics[width=1in,height=1.25in,clip,keepaspectratio]{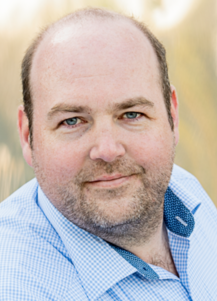}}]{Frank Diermeyer} ~holds the Diploma and Ph.D. degrees in mechanical engineering from the Technical University of Munich (TUM), Germany. Since 2008, he has been a Senior Engineer with the Chair of Automotive Technology, and the Leader of the Safe Operation Research Group within the Autonomous Vehicle Lab. His research interests include teleoperated driving, human–machine interaction, and safety validation.
\end{IEEEbiography}

\end{document}